\documentclass{article}

\PassOptionsToPackage{numbers, compress}{natbib}
 \usepackage[preprint]{neurips_2026}

\usepackage[utf8]{inputenc} % allow utf-8 input
\usepackage[T1]{fontenc}    % use 8-bit T1 fonts
\usepackage{hyperref}       % hyperlinks
\usepackage{url}            % simple URL typesetting
\usepackage{booktabs}       % professional-quality tables
\usepackage{amsfonts}       % blackboard math symbols
\usepackage{nicefrac}       % compact symbols for 1/2, etc.
\usepackage{microtype}      % microtypography
\usepackage[dvipsnames]{xcolor}         % colors
\usepackage{amsmath}
\usepackage{amsthm}
\usepackage{amssymb}
\usepackage[normalem]{ulem}
\usepackage{algorithm}
\usepackage{algpseudocode}
\usepackage{graphicx}
\usepackage{subcaption}
\usepackage{enumitem}
\usepackage{makecell}
\usepackage{tabularx}
\usepackage{pifont}
\newcommand{\cmark}{\ding{51}}%
\newcommand{\xmark}{\ding{55}}%

\theoremstyle{definition}
\newtheorem{definition}{Definition}[section]

\newtheorem{proposition}{Proposition}[section]

\theoremstyle{remark}

\DeclareMathOperator*{\argmax}{arg\,max}

\title{Propose to Learn, Learn to Propose: Evaluability-Aware Assistance under Bounded Rationality}

\author{%
Yifan Zhu$^{1,2}$ \qquad Sammie Katt$^{1,2}$ \qquad Samuel Kaski$^{1,2,3}$\\
$^1$ELLIS Institute Finland \\ 
$^2$Department of Computer Science, Aalto University, Finland \\
$^3$Department of Computer Science, University of Manchester, United Kingdom\\
\texttt{\{yifan.zhu, sammie.katt, samuel.kaski\}@aalto.fi}
}
\begin{document}

\maketitle

\begin{abstract}
AI assistants often collaborate by proposing candidate edits, plans, or designs that users evaluate before adoption.
Existing assistance methods focus on proposal quality or user-goal inference, often assuming that the user can reliably evaluate any proposal, which can fail in practice because of bounded rationality.
We study evaluability-aware proposal planning, where proposals serve both as task interventions and as probes for learning latent preferences and evaluation constraints, where the resulting belief updates then guide later proposals.
We formalise this setting as ProSE, a hidden-parameter sequential assistance problem, and instantiate it with a KL-regularised bounded-rational binary response model in which acceptance trades off value gain against a distance-dependent evaluability penalty.
Analysing the planning consequence of this likelihood reveals that likely accepted proposals and informative probes need not coincide, which explains why planners that only pursue acceptance systematically underperform.
We operationalise ProSE with \textsc{ProSE-Plan}, a depth-2 Bayes-adaptive planner that scores proposals by possible responses and response-induced posterior beliefs.
In controlled graph simulations, \textsc{ProSE-Plan} improves over evaluability-unaware and myopic baselines when evaluation cost is the bottleneck, and a probe-commit ablation confirms that our approach selects informative proposals that simpler methods miss.
Our results thus identify user evaluability as a planning-relevant dimension of AI assistance, complementary to generation quality and preference inference.
\end{abstract}

\section{Introduction}
In recent years, a proposal-evaluation paradigm has been widely applied in AI-assisted decision-making, where the assistant suggests a proposal, and the user decides whether to incorporate it through evaluation.
This is common in coding, writing, and design tasks, where assistance should preserve user control by proposing evaluable next states rather than acting autonomously.

Effective assistance in this paradigm requires more than generating high-quality proposals.
Existing assistance primarily models what the user wants, or how to act under uncertainty about the user's goals~\citep{fern2014decision,hadfield2016cooperative,ccelikok2022best,depeuter2023zero,laidlaw2025assistancezero}.
This addresses a central aspect of assistance, but neglects whether the user can evaluate a particular proposal from the current state.
This matters because human decision-making is bounded by cognitive and computational constraints~\citep{Simon1955ABM,Simon1956RationalCA,lewis2014computational,oulasvirta2022computational,howes2023towards}.
A sweeping code refactor may improve the codebase in principle, yet be too costly to inspect, while a smaller patch may be less globally optimal, but easier to verify, and therefore could be useful.
Empirical work on AI-assisted programming has exposed consistent concerns that users can struggle with proposals that are too complex, and excessive AI information can degrade rather than improve decision quality~\citep{fakhoury2024llm,alami2025human}.
While prior work has addressed aspects of this problem, such as bounded-rational user modelling~\citep{ccelikok2022best,depeuter2023zero}, query difficulty in preference learning~\citep{biyik2020asking,feng2025comparing}, and display timing~\citep{mozannar2024when} (discussed further in Appendix~\ref{sec:related_work}), how evaluability shapes the user's response and what the assistant can learn from it has received little attention in sequential assistance.

We formalise this problem as \textbf{pro}posal-based \textbf{s}equential assistanc\textbf{e} (ProSE).
In ProSE, each proposal has a dual role: it is a candidate intervention on the task and an observation probe for learning about the user.
The assistant must model the user's response both to predict what will happen if a proposal is shown and to update its belief about the user's latent parameters.
A high-value proposal may be rejected because it is difficult to evaluate, while a more moderate proposal may produce a response that improves future assistance.
Thus, the assistant should reason about how possible responses change future proposal choice, rather than only maximising immediate acceptance.

Within ProSE, we instantiate a tractable binary response model motivated by information-theoretic bounded rationality~\citep{ortega2013thermodynamics,ortega2015information}.
The response likelihood trades off user-value gain against a distance-dependent evaluability penalty, so the same proposal can be rejected either because it is low-value or too costly to evaluate from the current state.
We then analyse its structural properties and consequences for proposal planning.
The model induces an \emph{acceptance frontier} as the response decision boundary, and an \emph{information frontier}, characterising which proposal distances are informative about the user's evaluability parameter.
Our main result shows that the most informative proposal about evaluability can lie on the expected-rejection side of the acceptance frontier.
Thus, likely-to-be-accepted proposals and informative proposals need not coincide: rejection can be useful evidence for future assistance, not merely failed assistance.

We operationalise the framework with \textsc{ProSE-Plan}, a depth-2 Bayes-adaptive planner that scores proposals by possible responses and response-induced posterior beliefs.
We evaluate it in two controlled graph tasks: a branching corridor for end-to-end performance when high-value proposals are difficult to evaluate, and a probe-commit task isolating response-dependent belief updates.
The results show that evaluability-aware planning helps when evaluation cost is the bottleneck, and that anticipating what a response reveals about the user leads to better subsequent proposals.
Our contributions are to formulate evaluability-aware proposal planning, instantiate it with a bounded-rational response model and frontier analysis, and validate a minimal Bayes-adaptive planner in controlled simulations.
Together, these results identify user evaluability as a planning-relevant dimension of AI assistance, complementary to generation quality and preference inference.

\section{Preliminaries} \label{sec:preliminaries}

Our work utilises two important concepts: Bayesian decision-making under latent model uncertainty and information-theoretic bounded rationality.
This section briefly introduces both.

\textbf{Bayesian decision-making.}
Sequential decision-making is typically formalised as a Markov decision process (MDP) $(\mathcal S,\mathcal A,\mathcal T,\mathcal R,\gamma)$, where $\mathcal S$ is a state space, $\mathcal A$ is an action space, $\mathcal{T}(s'\mid s,a): \mathcal S \times \mathcal A \to \Delta(\mathcal S)$ is the transition dynamics, $\mathcal R(s,a,s'): \mathcal S \times \mathcal A \times \mathcal S \to \mathbb R$ is the reward function, and $\gamma\in[0,1]$ is a discount factor.
The objective is to find a policy $\pi: \mathcal{S} \to \Delta(\mathcal{A})$ that maximises expected return:  $\pi^\star\in\argmax_{\pi}\mathbb E_{\tau\sim(\pi,\mathcal{T})}\left[\sum_{t=0}^{T-1}\gamma^t \mathcal{R}(s_t,a_t,s_{t+1})\right]$, where $a_t\sim\pi(\cdot\mid s_t)$, $s_{t+1}\sim \mathcal{T}(\cdot\mid s_t,a_t)$, and $T$ is the planning horizon~\cite{Sutton2018}.

In many settings, the transition or reward model is not known \textit{a priori}, often parameterised by latent parameters $z$ and learned from experiences, which is costly in real-world applications.
The Bayesian alternative is to assume a prior $p_0(z)$ and maintain a belief $b_t(z)=p(z\mid h_t)$ over the latent parameter given the interaction history $h_t = (s_0, a_0, \dots, a_{t-1}, s_t)$, so that the agent is able to make decisions that are optimal with respect to evolving belief over the world.
Specifically, after observing a transition $(s_t,a_t,s_{t+1})$, the belief is updated by Bayes' rule:
\begin{equation}
b_{t+1}(z)=p(z\mid h_t,a_t,s_{t+1})\propto \mathcal{T}_z(s_{t+1}\mid s_t,a_t)b_t(z).
\label{eq:bayes_update_latent_mdp}
\end{equation}
The optimal solution is then the policy that optimises future reward with respect to this belief.
The Bayes-adaptive MDP (BA-MDP) formalises this idea by casting planning under latent model uncertainty in an augmented belief state $(s,b)$~\cite{duff2002optimal}.
Typical solutions to (BA-)MDPs compute and maximise the Q-value: the expected value of taking an action in a state, which can be computed recursively with the Bellman equation with $h$ steps of lookahead remaining:
\begin{equation}
\label{eq:bamdp-q}
Q_h^\star(s,b,a)
=
\int_z b(z)
\sum_{s'} \mathcal{T}_z(s'\mid s,a)
\left[
\mathcal{R}_z(s,a,s')
+
\gamma \max_{a'\in\mathcal A}Q_{h-1}^\star(s',b^{a,s'},a')
\right]
dz,
\end{equation}
where $Q_0(\cdot)$ is typically defined as 0, $b^{a,s'}(z)$ is the posterior belief after observing $s'$.

\textbf{Information-Theoretic Bounded Rationality.}
Bounded rationality refers to the observation that agents operate under internal cognitive and computational constraints that prevent them from fully optimising their decisions~\citep{Simon1955ABM, Simon1956RationalCA}.
The computational-rationality perspective treats such deviations from perfect rationality as a result of agents making the best use of their limited cognitive resources under subjective utility~\citep{lewis2014computational, gershman2015computational,oulasvirta2022computational}.
In particular, information-theoretic bounded rationality~\citep{ortega2013thermodynamics, ortega2015information} formalises user behaviour as solving a one-shot KL-regularized decision problem, where the probability of picking an action $a \in \mathcal{A}$ is described as maximizing the trade-off between utility $U:\mathcal A\to\mathbb R$ and information cost in the form of deviating from the prior policy $q\in\Delta(\mathcal A)$: %\skattInline{unclear what prior policy here could be (``the thing a user would do if not thinking''?)}
\begin{equation}\label{eq:information-theoretic-bounded-rationality}
  p^{\mathrm{ITBR}} = \argmax_{p \in \Delta(A)} \Big[\sum_{a \in \mathcal{A}} p(a) U(a) - \frac{1}{\kappa} D_\mathrm{KL}(p \mid\mid q)\Big]
  %p^{\mathrm{ITBR}} = \argmax_{p \in \Delta(A)} \Big[\sum_{a \in \mathcal{A}} p(a) U(a) - \frac{1}{\kappa} \underbrace{\sum_{a \in \mathcal{A}}p(a) \log \frac{p(a)}{q(a)}}_{D_\mathrm{KL}(p \mid\mid q)}\Big]
\end{equation}
where $\kappa > 0$ is an inverse-temperature parameter that controls the user's cognitive effort.
A larger $\kappa$ indicates the user invests more effort in deviating from the default to pursue higher utility, while $\kappa \to 0$ recovers the default action regardless of utility.
Since the objective in Eq.~\eqref{eq:information-theoretic-bounded-rationality} is strictly concave in $p$ whenever $q(a) > 0$ for all $a$, it admits a unique closed-form solution (Appendix~\ref{sec:gibbs_derivation} or~\cite{ortega2015information} for details).
Solving the simplex-constrained optimization gives the Gibbs policy:
\begin{equation}\label{eq:itbr-gibbs}
  p^{\mathrm{ITBR}}(a) \propto q(a) \exp \big(\kappa U(a) \big).
\end{equation}
Eq.~\eqref{eq:itbr-gibbs} shows that the optimal action is the default distribution $q$ tilted toward higher-utility actions, with the degree of tilting controlled by $\kappa$.
%This holds for any finite response space $\mathcal{Y}$, any utility function $U$, and any default distribution $q_\theta$ with full support.
%The KL-regularized framework therefore determines the \emph{form} of a principled response likelihood, while the ProSE-specific components of the model, such as decision utility and how the default response depends on the proposal, remain as modelling choices to instantiate proposal-dependent evaluability, which we specify next.

\section{Problem Setup: Proposal-based Sequential Assistance}\label{sec2:setup}
We study a sequential design task family where the user iteratively revises the current design (state) toward a satisfactory outcome.
In this interaction paradigm, the assistant proposes candidate next states or edits, which the user then evaluates and responds to (e.g., accept or reject), and the realised next state is then determined by the current state, the proposal, and the user's response.
The assistant's primary goal is to help the user achieve a better final artefact through proposals, and the main difficulty is that good proposals depend on the user's likelihood of accepting them.
As we've motivated above, this likelihood depends not only on its quality, i.e., how much it aligns with the user's preferences, but also on whether a user can (and wants to) evaluate it.
In this section, we formalise this problem setting as \textbf{Pro}posal-based \textbf{S}equential assistanc\textbf{E} (ProSE).
% \subsection{Interaction Formulation}
\begin{definition}[ProSE process]\label{def:prose}
A ProSE process consists of
a task state space $\mathcal S$,
a candidate proposal space $\mathcal{C}$ with $\mathcal{C}(s) \subseteq \mathcal{S}$,
a user response space $\mathcal Y$,
a latent user parameter space $\mathcal Z = \Phi \times \Theta$ with prior $p_0$ characterising preferences ($\Phi$) and evaluability ($\Theta$) respectively,
a user response model family $P(y\mid s,\tilde{s}, z): \mathcal{S} \times \mathcal{C}(s) \times \mathcal{Z} \to \Delta(\mathcal{Y})$,
response-mediated transition dynamics $\mathcal{T}^y(s'\mid s,\tilde{s},y): \mathcal{S} \times \mathcal{C}(s) \times \mathcal{Y} \to \Delta(\mathcal{S})$,
and a family of user value functions $\mathcal V_\Phi=\{V_\phi: \mathcal{S} \to \mathbb{R} \}_{\phi \in \Phi}$.
\end{definition}

In ProSE, the assistant's action space is a set of proposals $\mathcal{C}(s)$, which we will refer to as $\tilde s$ (as opposed to action $a$).
At each time step, given a selected proposal $\tilde{s}$, the user responds according to $y \sim P(\cdot \mid s, \tilde s, z)$, and the task state evolves according to $s'\sim\mathcal T^y(\cdot\mid s,\tilde{s},y)$.
In design tasks, the dynamics are typically deterministic given user actions, which we encode with the deterministic transition function $\mathcal{T}^y(s' \mid s, \tilde s, y)$.
For example, the user's reply $y$ could be to ignore the proposal and apply some edit themself to change the state.

ProSE is a specific instantiation of the Bayes-adaptive MDP.
Specifically, given user parameters $z$, the problem can be fully formalised and solved as an MDP, with two notable details:
(1) the transition model is defined by the user model, and (ii) the reward function is zero $\mathcal{R}(s_t, a, s_{t+1}) = 0$ except for the last state $s_T$, of which the value is determined by the user $\mathcal{R}(\cdot, \cdot, s_T) = V_\phi(s_T)$.
Of course, the user parameters are not known.
This is resolved by assuming a prior $p_0(z) = p_0(\theta, \phi)$ over the parameters and casting the problem as a Bayes-adaptive MDP as discussed in Section~\ref{sec:preliminaries}.
Here, the transitions are fully specified by $z$ and the belief over $\mathcal{T}$ can be summarized with a belief over $z$: 
\begin{equation}\label{eq:prose-belief-update}
    b_t(z) = p(z \mid h_t) \propto P(y_{t-1} \mid s_{t-1}, \tilde{s}_{t-1}, z) b_{t-1}(z).
\end{equation}
The optimal solution also follows from the Bayes-adaptive formulation (Eq.~\ref{eq:bamdp-q}), with the notable adaptation that the value of the last step is now determined by the user: $Q_0 = V_\phi$.

\section{Bounded-Rational Binary User Response Model}\label{sec:response_model}
In Section~\ref{sec2:setup}, we reduced the assistant's problem to belief-state planning over a response likelihood $P(y\mid s,\tilde s, z)$.
Here, we derive a concrete \emph{binary} model for users in design tasks where their response is either to accept or reject the AI's proposal ($\mathcal Y=\{\mathrm{acc},\mathrm{rej}\}$).
This results in a process that either transitions to the proposed state $\tilde s$, if the user accepts, or leaves the state $s$ unchanged if rejected: 
$$T^y(s' \mid s,\tilde{s}, \text{acc}) = \mathbf{1}\{s'=\tilde{s}\} \text{ and } T^y(s' \mid s,\tilde{s}, \text{rej}) = \mathbf{1}\{s'=s\}.$$
This setting models significant parts of interactions with LLMs in coding and writing, while giving a minimal model in which evaluability can affect both task progress and user-model learning, demonstrating interesting behaviour and insights as we show in the analysis in Section~\ref{sec:user-model-analysis}.

\subsection{A Tractable Evaluability-Aware Response Model}\label{sec:binary_model} 
Here, we derive the binary user model from first principles.
By applying Eq.~\eqref{eq:itbr-gibbs} to proposal responses in ProSE, we have $P(y\mid s,\tilde{s},z) \propto q_\theta(y\mid s,\tilde{s})\exp\left(\kappa U(y;s,\tilde{s},\phi)\right)$, where $q_\theta$ is the user's low-effort default response, and $U(y;s,\tilde{s},\phi)$ is the utility of response $y$.
For binary responses, this gives 
\begin{equation}
P(\mathrm{acc}\mid s,\tilde{s},z)=\sigma\left(\kappa \Delta U_\phi(s,\tilde{s})+\lambda_\theta(s,\tilde{s})\right),
\label{eq:binary-response-sigmoid}
\end{equation}
where $\Delta U_\phi(s,\tilde{s})=U(\mathrm{acc};s,\tilde{s},\phi)-U(\mathrm{rej};s,\tilde{s},\phi)$ and $\lambda_\theta(s,\tilde{s}) = \log \frac{q_\theta(\mathrm{acc}\mid s,\tilde{s})}{q_\theta(\mathrm{rej}\mid s,\tilde{s})}$ (see Appendix~\ref{app:binary-logistic} for details).
The response utility $U$ is induced by the user's latent value function.
For a general response $y$, define $U(y;s,\tilde{s},\phi)=\mathbb E_{s'\sim \mathcal T^y(\cdot\mid s,\tilde{s},y)}[V_\phi(s')]$.
Under the deterministic binary transition above, $U(\mathrm{acc};s,\tilde{s},\phi)=V_\phi(\tilde{s})$ and $U(\mathrm{rej};s,\tilde{s},\phi)=V_\phi(s)$.
The utility difference therefore becomes the value gain, quantifying the difference in user preference between the proposed and current state: % according to the user's preferences:
\begin{equation}\label{eq:value_gain}
\Delta U_\phi(s,\tilde{s})=V_\phi(\tilde{s})-V_\phi(s)=:\Delta V_\phi(s,\tilde{s}).
\end{equation}

As for the default response log-odds $\lambda_\theta(s,\tilde{s})$, instead of assuming or estimating an arbitrary default policy, we parameterise it directly to represent evaluability burden.
Here, we make a modelling choice that the default log-odds decreases with proposal distance, to capture proposal-dependent evaluability, where a proposal farther from the current state requires more cognitive effort to evaluate and therefore faces a stronger default toward rejection.
This choice is motivated by evidence that status-quo bias scales with the magnitude of the proposed change~\citep{Zeckhauser1988Status} and that cognitive load increases with the complexity of the evaluation task~\citep{Sweller1988Cognitive}.
Thus, the default log-odds is instantiated as 
\begin{equation}\label{eq:default_logodds}
\lambda_\theta(s,\tilde{s})
= -\alpha\,g\bigl(d(s,\tilde{s})\bigr),
\end{equation}
where $d(s,\tilde{s})\ge 0$ measures proposal distance on the state space, $\alpha\ge 0$ controls the overall strength of the distance penalty, and $g$ is a non-decreasing function with $g(0)=0$ that maps distance to evaluation burden.
We use a transform $g$ rather than raw distance because evaluation difficulty need not scale proportionally with distance~\citep{Zeckhauser1988Status}: assessing a small edit may be easy, while a moderately larger change can be disproportionately harder to evaluate.

Substituting Eqs.~\eqref{eq:value_gain} and \eqref{eq:default_logodds} into Eq.~\eqref{eq:binary-response-sigmoid} gives a tractable predictive likelihood:
\begin{equation}
  \label{eq:response_model}
  P(\mathrm{acc}\mid s,\tilde{s},z)
  =\sigma(\kappa[\Delta V_\phi(s,\tilde{s})-\rho\,g(d(s,\tilde{s}))]),
  \end{equation}
where $\rho:=\alpha/\kappa$.
This instantiates the user parameters $z=(\phi,\theta)$ from Section~\ref{sec2:setup} as $z=(\phi,\rho,\kappa)$.
In this response model, the preference parameter $\phi$ determines the user's preference over states, while the evaluability slope $\rho\ge0$ determines how much proposal distance shifts the acceptance threshold, and the response sharpness $\kappa>0$ controls how decisively the user responds to the value-cost difference.

\subsection{User Model Analysis: Acceptance and Information Frontiers} \label{sec:user-model-analysis}
We now analyse the response model in Eq.~\eqref{eq:response_model} to identify the structural properties and consequences it induces for proposal planning.
We first derive the \emph{acceptance frontier}, which characterises when a proposal is more likely to be accepted than rejected, and shows that distance raises the value gain required for acceptance.
We then derive an \emph{information frontier}, which characterises which proposal distances make the response most informative about the user's evaluability slope $\rho$.
These frontiers give two planning lessons.
First, likely-to-be-accepted proposals and informative proposals need not coincide, so maximising immediate acceptance can miss useful probes.
Second, the most informative proposals about evaluability can lie on the expected-rejection side of the acceptance frontier, rather than on the frontier where the response uncertainty is maximal.
This indicates that rejection is not always merely failed assistance, but can be an informative response that improves future proposal choice through user-model learning.
The theoretical results thus motivate planning over response-dependent belief updates rather than myopic acceptance maximisation.

\begin{figure}[t]
\centering
\begin{subfigure}[t]{0.45\textwidth}
\centering
\includegraphics[width=\linewidth]{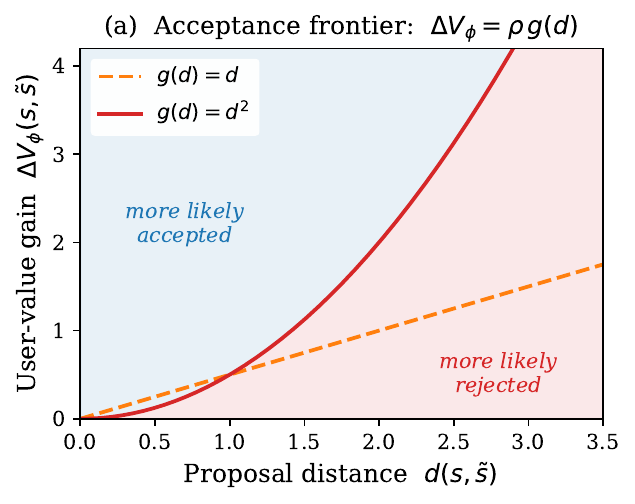}
\end{subfigure}
\hfill
\begin{subfigure}[t]{0.49\textwidth}
\centering
\includegraphics[width=\linewidth]{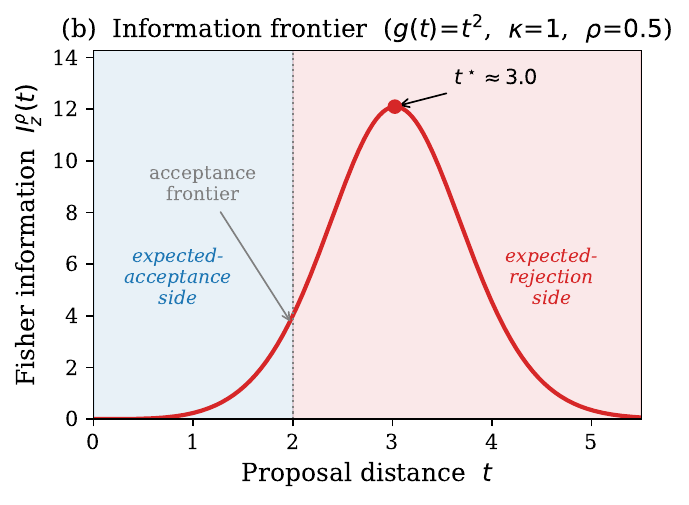}
\end{subfigure}
\caption{%
\textbf{Acceptance and information frontiers.
}
(a) The acceptance frontier $\Delta V_\phi=\rho\,g(d)$ separates likely acceptance and rejection.
(b) With $g(t)=t^2$, $\kappa=1$, $\rho=0.5$, and $\Delta V_\phi(t)=t$, Fisher information about $\rho$ peaks beyond the acceptance frontier, where rejection is more probable. 
}\label{fig:frontiers}
\end{figure}

\textbf{Acceptance Frontier.}
In Eq.~\eqref{eq:response_model}, acceptance is more likely when the user-value gain $\Delta V_\phi$ exceeds the evaluability penalty $\rho\,g(d)$.%, and less likely otherwise.
The boundary where these two terms balance,
\begin{equation}
\label{eq:acceptance_frontier}
\Delta V_\phi(s,\tilde{s})=\rho\,g(d(s,\tilde{s})),
\end{equation}
defines the \emph{acceptance frontier}.
It is the curve in the $(\text{distance},\,\text{value gain})$ plane separating proposals that are more likely to be accepted from those that are more likely to be rejected, as illustrated in Figure~\ref{fig:frontiers}(a).
The frontier captures the immediate effect of evaluability: farther proposals require larger value gains to remain acceptable.

\textbf{Information Frontier.}
The acceptance frontier characterises immediate acceptance.
For sequential proposal planning, the assistant also needs to know which responses are useful for learning the latent user parameters.
We therefore analyse which proposal distances make a single accept/reject response informative about the evaluability slope $\rho$.

To isolate the structural effect of proposal distance, we analyse the response likelihood conditional on a latent user-parameter $z=(\phi,\rho,\kappa)$.
This does not assume that the assistant knows $z$, but characterises the likelihood term that the assistant later integrates under its posterior belief.
Consider a one-dimensional proposal path $\{\tilde{s}(t)\}_{t\ge0}$ from the current state $s$, parameterised by distance so that $\tilde{s}(0)=s$ and $d(s,\tilde{s}(t))=t$.
Along this path, define
\begin{equation}
\label{eq:path_logit}
\eta_z(t)=\kappa[\Delta V_\phi(t)-\rho g(t)],
\qquad
\Delta V_\phi(t):=V_\phi(\tilde{s}(t))-V_\phi(s),
\end{equation}
and let $p_z(t)=\sigma(\eta_z(t))$ be the acceptance probability.

Since the response is binary, observing the user's response gives one Bernoulli observation with parameter $p_z(t)$.
We measure how informative this observation is about $\rho$ using Fisher information, a local sensitivity measure of the likelihood with respect to a parameter~\citep{Chaloner1995BayesianED, MacKay1992info}.
For the Bernoulli likelihood induced by Eq.~\eqref{eq:response_model}, the Fisher information about $\rho$ is (see Appendix~\ref{app:conditional-information-frontier} for details)
\begin{equation}
\label{eq:rho_fisher_information}
I_z^\rho(t)=\kappa^2 g(t)^2p_z(t)(1-p_z(t)).
\end{equation}
Eq.~\eqref{eq:rho_fisher_information} contains two competing factors.
The term $g(t)^2$ grows with proposal distance and captures sensitivity to $\rho$: when a proposal is very close to the current state, changing $\rho$ has little effect on the response probability.
The term $p_z(t)(1-p_z(t))$ is the Bernoulli response variance, which is largest when acceptance and rejection are equally likely and vanishes when the response is almost deterministic.
Thus, very local proposals are weakly informative because they are insensitive to $\rho$, while very distant proposals are weakly informative because they are almost surely rejected.
Therefore, Eq.~\eqref{eq:rho_fisher_information} indicates that information about $\rho$ must peak at some intermediate distance.

Now, we identify on which side of the acceptance frontier the peak lies with the following proposition.

\begin{proposition}[Information frontier conditional on user parameters]
\label{prop:conditional_information_frontier}
For any latent user-parameter vector $z=(\phi,\rho,\kappa)$ with $\rho>0$ and $\kappa>0$, assume $\Delta V_\phi(t)$ is continuously differentiable, $\Delta V_\phi(t)=o(g(t))$ as $t\to\infty$, and $g$ satisfies $g(0)=0$, $g(t)>0$, $g'(t)>0$ for all $t>0$, and $g(t)\to\infty$.
Then $I_z^\rho(t)$ in Eq.~\eqref{eq:rho_fisher_information} satisfies:
(i)~$I_z^\rho(0)=0$;
(ii)~$I_z^\rho(t)\to0$ as $t\to\infty$;
(iii)~$I_z^\rho(t)$ attains a global maximum at some $t^\star\in(0,\infty)$; and
(iv)~every global maximiser satisfies $p_z(t^\star)<1/2$.
\end{proposition}
Proof provided in Appendix~\ref{app:information-frontier-proof}. Figure~\ref{fig:frontiers}(b) illustrates this effect for $g(t)=t^2$, $\kappa=1$, $\rho=0.5$, and $\Delta V_\phi(t)=t$: the information maximum lies beyond the acceptance frontier, where rejection is more likely than acceptance.
At the frontier, response uncertainty is maximal, but the sensitivity factor $g(t)^2$ is still increasing, pushing the information maximum to the expected-rejection side.

This separates two notions that a myopic assistant can conflate.
A proposal can be easy to accept but uninformative about evaluability, or likely to be rejected but useful for locating the user's evaluability boundary.
The information frontier is therefore not a new planning objective, but a diagnostic of the response model: it shows why a sequential assistant should reason about how each possible response changes its belief over the user, rather than only maximising immediate acceptance.
This is the motivation for the response-dependent belief updates used by our planner in Section~\ref{sec:method}.

\section{\textsc{ProSE-Plan}: Evaluability-Aware Proposal Planning}\label{sec:method}
The ProSE problem in Section~\ref{sec2:setup} is a Bayes-adaptive belief-state planning problem which, in principle, can be solved optimally with standard RL solutions once the response likelihood, transition model, and value family are specified.
Our goal in this section is not to contribute a novel \emph{general} solution, but to introduce a minimal planner, \textsc{ProSE-Plan}, to analyse solutions to our problem setting.

\textbf{Belief Tracking}
We track the posterior belief over the latent user parameters $p(z)$ with $z = (\phi, \rho, \kappa)$ according to Eq.~\ref{eq:prose-belief-update}.
For tractability, we represent $z$ on a finite grid and maintain the posterior exactly on that grid. 
More sophisticated Bayesian approaches are available for large parameter spaces.

\textbf{Proposal Selection}
We propose a depth-2 look-ahead search which scores each proposal by the 2-step Q-value $Q_h$ (recall Eq.~\eqref{eq:bamdp-q} in Section~\ref{sec2:setup}).
This means, for every possible proposal $\tilde s \in \mathcal{C}(s)$, we compute the next state likelihood given our belief of the user accepting it:
\begin{equation}\label{eq:planner-transition-likelihood}
    p(s' \mid b, s, \tilde s) = \sum_y \mathbb{E}_{z \sim b} \big[ \mathcal{T}^y(s' \mid s, \tilde s, y) P(y \mid s, \tilde s, z) \big]
\end{equation}
Then, for each of those hypothetical state transitions, the corresponding new hypothetical belief $b_{s'}$ is computed (Alg~\ref{alg:prose-plan} line~\ref{eq:prose-belief-update} in Appendix~\ref{app:planner-details}), and this \emph{response-dependent belief update} is repeated for a second step to compute the likelihood for all hypothetical future trajectories of depth 2.
Then, the Q-value of each node in the graph is computed, where the value of the last step is the expectation over the user preferences: $Q_0(b, s) = \mathbb{E}_{\phi \sim b} \big[ V_\phi(s) \big]$ (recall Eq.~\ref{eq:bamdp-q}).
Ultimately, the proposal with the highest expected value $Q_2$ is chosen.
Deeper lookahead and Monte Carlo tree search~\citep{silver2010monte} are promising improvements, but this solution is sufficient for our purposes.

Under the binary deterministic transition in Section~\ref{sec:binary_model}, the expectations over future states collapse to the accepted or rejected next state.
With maximum candidate-set size $|\mathcal C|$, response-space size $|\mathcal Y|$, and grid size $|\mathcal Z_{\mathrm{grid}}|$, exhaustive depth-2 evaluation costs $O(|\mathcal C|^2|\mathcal Y|^2|\mathcal Z_{\mathrm{grid}}|)$ per planning step.

\begin{figure}[t]
  \centering
  \begin{subfigure}[t]{0.55\columnwidth}
    \centering
    \includegraphics[width=\linewidth]{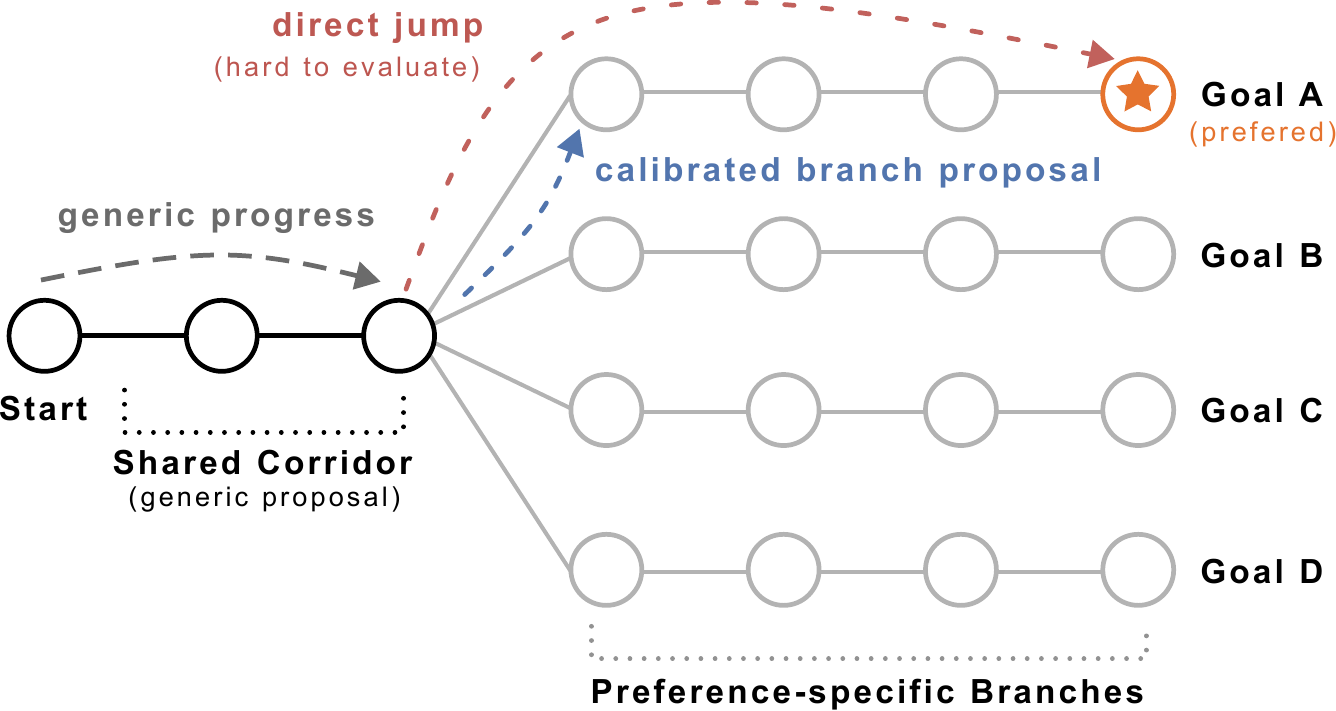}
    \caption{\textbf{Branching corridor.}}
    \label{fig:corridor_illustration}
  \end{subfigure}
  \hfill
  \begin{subfigure}[t]{0.3\columnwidth}
    \centering
    \includegraphics[width=\linewidth]{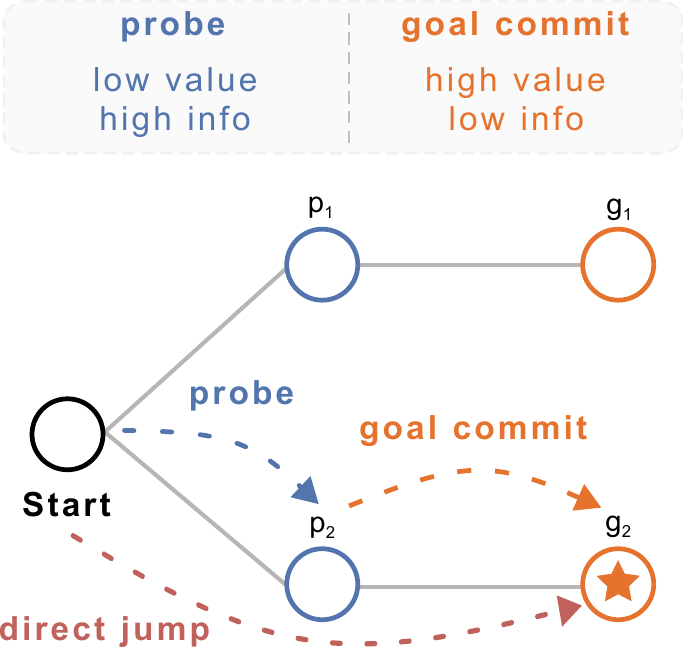}
    \caption{\textbf{Probe-commit setting.}}
    \label{fig:isolation_illustration}
  \end{subfigure}
  \caption{
    Illustrations for experiment settings.
  }
  \label{fig:env_illustration}
  \vspace{-4pt} 
\end{figure}

\section{Experiments}\label{sec:experiment}
In this section, we use controlled graph-based simulations to empirically test whether the response model from Section~\ref{sec:response_model} has planning consequences in finite-horizon sequential tasks. 
Specifically, we ask the following two research questions:
\begin{enumerate}[label=\textbf{Q\arabic*},leftmargin=*,nosep]
  \item \emph{End-to-end payoff.}
  Does evaluability-aware planning improve final outcomes over methods that ignore evaluability, user-specific evaluability, or non-myopic proposal planning?
  \item \emph{Mechanism ablation.}
  Does the gain of \textsc{ProSE-Plan} come from depth-2 lookahead alone, or from response-dependent belief updates inside the lookahead tree? 
\end{enumerate}
We evaluate \textbf{Q1} in a branching-corridor task and \textbf{Q2} in a minimal probe-commit task, and leave the main sensitivity analysis to Appendix~\ref{app:experiments}.

\subsection{Experimental Setup}\label{sec:exp_setup}
We evaluate \textsc{ProSE-Plan} in controlled finite-graph proposal tasks. 
In each task, states are graph nodes, the assistant proposes a candidate node $\tilde{s}$, and evaluation difficulty is represented by $d(s,\tilde{s})$, the unweighted shortest-path distance between two nodes. 
User responses follow the bounded-rational response model in Eq.~\eqref{eq:response_model}, accepting a proposal moves the state to $\tilde{s}$, while rejecting it leaves the state unchanged. 
The task terminates if the user reaches the preferred goal or the horizon is reached.
The latent user parameter $z=(\phi,\rho,\kappa)$ determines the user's preferred goal, evaluability cost, and response sharpness. 
To avoid confounding the evaluation of planning with approximate inference errors, the assistant maintains an exact discrete posterior over $z$. 
We report success rate, defined as the fraction of episodes ending at the user's preferred goal state, and mean terminal user value $V_\phi(s_T)$, averaged over 200 random seeds per condition.

We consider the following baselines and ablation. 
\textsc{Random} selects proposal uniformly.
\textsc{Value-greedy} ignores evaluability and selects the proposal with the highest value.
Inspired by the satisficing alignment principle~\cite{chehade2025bounded}, \textsc{Threshold} greedily proposes from a candidate set constrained by a distance threshold.
Both \textsc{Population-myopic} and \textsc{Personalised-myopic} are myopic planners, while the former uses population-level evaluability parameters and the latter infers the full latent user state.
\textsc{Belief-frozen depth-2} is a mechanism ablation, which has the same depth-2 horizon as \textsc{ProSE-Plan}, but evaluates future proposals under the current belief rather than the posterior induced by hypothetical first-step responses. 
\textsc{Oracle depth-2} knows the true user parameter and serves as an upper-bound reference. 
All methods infer $\phi$ except \textsc{Random} and \textsc{Oracle depth-2}.
Full descriptions are provided in Appendix~\ref{app:baselines}.

\begin{figure}[t]
  \centering
  \begin{subfigure}[t]{0.58\columnwidth}
    \centering
    \includegraphics[width=\linewidth]{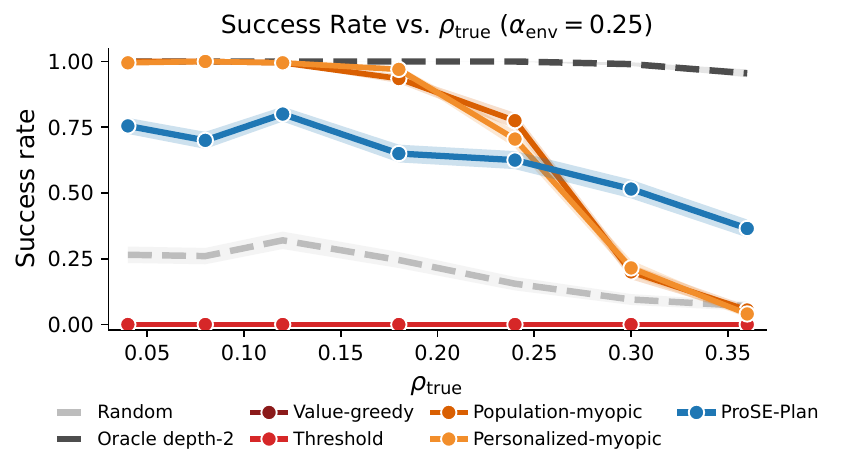}
    \caption{Success vs. $\rho_{\mathrm{true}}$.}
    \label{fig:corridor_success}
  \end{subfigure}
  \hfill
  \begin{subfigure}[t]{0.4\columnwidth}
    \centering
    \includegraphics[width=\linewidth]{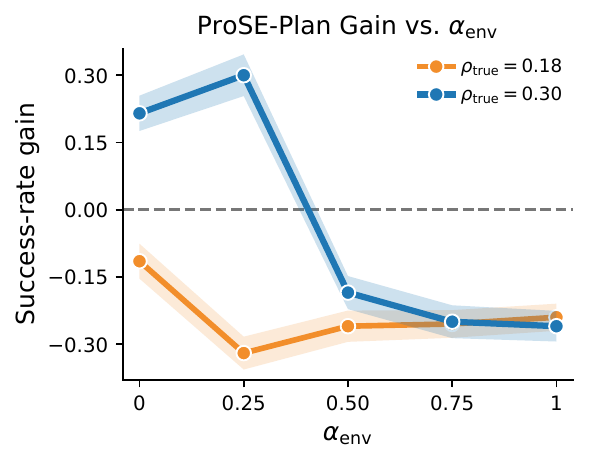}
    \caption{Gain vs. $\alpha_{\mathrm{env}}$.}
    \label{fig:corridor_advantage}
  \end{subfigure}
  \caption{
  Branching-corridor results, mean $\pm$ standard error over 200 seeds.
  (a) Success rate versus true evaluability cost at $\alpha_{\mathrm{env}}=0.25$.
  (b) Success-rate gain of \textsc{ProSE-Plan} over the stronger myopic baseline.
  Gains appear only when the evaluation cost is high and shared-corridor value is low.
  }
  \label{fig:corridor}
\end{figure}

\subsection{End-to-End Payoff in a Branching Corridor}\label{sec:exp_corridor}
\iffalse
We construct a branching-corridor task to test whether evaluability-aware planning improves end-to-end outcomes when the assistant must trade off easy generic progress against harder but more personalised proposals. 
As illustrated in Figure~\ref{fig:env_illustration}a, the environment consists of a start state, a shared corridor, and $K$ preference-specific branches. 
The latent preference $\phi \in \{1,\dots,K\}$ specifies which branch contains the user's true goal. 
We design $V_\phi(s)$ such that later shared-corridor nodes have increasing generic value for all users, later nodes on the preferred branch have additional user-specific value, and later nodes on a wrong branch have lower value. 
This creates a proposal-level trade-off, where shared-corridor proposals are easy to evaluate and accept, but provide little information about $z$, while branch proposals are more informative and can yield larger value gains if aligned with the user, but are harder to evaluate and risk rejection when they target the wrong branch or jump too far.
Detailed environmental setup is provided in Appendix~\ref{app:corridor_full}.
\fi

We construct a branching-corridor task to test whether evaluability-aware planning improves end-to-end outcomes when the assistant must trade off easy generic progress against harder but more personalised proposals. 
The branching corridor (Figure~\ref{fig:env_illustration}a) is a parametric graph-based sequential task family, where the state space is an undirected tree with a start node $s_0$, a shared corridor $\{c_1, \ldots, c_C\}$, and $K$ branches of length $L$ attached at $c_C$, with branch nodes $\{b_{k,1}, \ldots, b_{k,L}\}$ for $k \in \{1, \ldots, K\}$.
The user has a hidden preferred branch $\phi \in \{1, \ldots, K\}$ with goal leaf $b_{\phi,L}$.
The user's value function family combines shared corridor and branch progress: $V_\phi(c_j) = \alpha_{\mathrm{env}}\, w_c\, j, V_\phi(b_{\phi,j}) = \alpha_{\mathrm{env}}\, w_c\, C + w_b\, j, V_\phi(b_{k,j}) = \alpha_{\mathrm{env}}\, w_c\, C - w_p\, j \;\text{for}\; k \neq \phi$, 
where $V_\phi(s_0) = 0$, $\alpha_{\mathrm{env}} \in [0,1]$  is a task-structure parameter controlling the relative value between common progress and personalised commitment.
At each step $t=0,\ldots,T-1$, the assistant proposes a node $\tilde{s}_t \neq s_t$, the user responds with $y_t \in \{\mathrm{acc}, \mathrm{rej}\}$ via Eq.~\eqref{eq:response_model} under latent $z = (\phi, \rho, \kappa)$, and the state moves to $\tilde{s}_t$ if $y_t = \mathrm{acc}$ and stays at $s_t$ otherwise.
We define success in this case as reaching $b_{\phi, L}$ within the horizon $T$.
In experiments, we use $K=4$, $C=2$, $L=4$, $T=5$, $(w_c, w_b, w_p) = (3, 2, 3)$ as defaults, consider a quadratic transform $g(d) = d^2$, and $\mathcal{C}(s)=\mathcal{S}\setminus\{s\}$.
More details are provided in Appendix~\ref{app:corridor_setup}.

\textbf{Main results (\textbf{Q1}).}
Figure~\ref{fig:corridor}(a) shows that \textsc{ProSE-Plan} improves end-to-end success when proposal evaluability becomes the bottleneck.
At low $\rho_{\mathrm{true}}$, distant proposals are easy to evaluate, so the strongest baselines, especially \textsc{Population-myopic} and \textsc{Personalised-myopic}, achieve near-ceiling success.
As $\rho_{\mathrm{true}}$ increases, these baselines collapse because they continue to propose distant branch states whose value is high under some preferences, but whose acceptance probability becomes low under high evaluation cost.
In contrast, \textsc{ProSE-Plan} degrades more gracefully and becomes the best non-oracle method in the high-cost setting by selecting closer, more evaluable proposals.
Table~\ref{tab:final_value} gives the default high-cost comparison at $\alpha_{\mathrm{env}}=0.25$, $\rho_{\mathrm{true}}=0.30$: \textsc{ProSE-Plan} reaches success $0.515$, more than twice \textsc{Personalised-myopic} ($0.215$) and \textsc{Population-myopic} ($0.200$).
Aggregated proposal behaviour and representative trajectories in Appendix~\ref{app:corridor_proposal_behaviour} and \ref{app:corridor_trajectories} show the corresponding behavioural pattern.
Thus, the branching-corridor result supports \textbf{Q1}: evaluability-aware proposal planning improves final outcomes when high-value personalised proposals are difficult to evaluate.

\textbf{Sensitivity to task structure.}
Figure~\ref{fig:corridor}(b) shows that the gain of \textsc{ProSE-Plan} depends on the task structure.
At high evaluation cost ($\rho_{\mathrm{true}}=0.30$), \textsc{ProSE-Plan} improves over the stronger myopic baseline when $\alpha_{\mathrm{env}}$ is small, with the largest gain at $\alpha_{\mathrm{env}}=0.25$.
As $\alpha_{\mathrm{env}}$ increases, shared-corridor states already provide enough value, so simple myopic baselines become competitive.
At lower evaluation cost ($\rho_{\mathrm{true}}=0.18$), the gain is consistently non-positive because distant proposals remain sufficiently evaluable.
Thus, evaluability-aware planning helps most when generic local suggestions are easy to accept but insufficient for reaching the user's intended outcome, while personalised high-value proposals are costly to evaluate.

\begin{table}[t]
\centering
\caption{%
Results of the Branching Corridor and Probe-Commit tasks.
Branching-Corridor uses $\alpha_{\mathrm{env}}=0.25$ and $\rho_{\mathrm{true}}=0.30$;
Probe-Commit uses $w_{\mathrm{probe,mismatch}}=-3$, $\rho_{\mathrm{true}}=0.5$, and $\kappa_{\mathrm{true}}=1.0$.
Values are reported as mean $\pm$ standard error over 200 seeds.}
\label{tab:final_value}
\small
\begin{tabular}{lccccc}
\toprule
  & \multicolumn{2}{c}{Branching Corridor} & \multicolumn{3}{c}{Probe-Commit} \\
\cmidrule(lr){2-3} \cmidrule(lr){4-6}
Method & $V_\phi(s_T)$ & Success rate & $V_\phi(s_T)$ & Success rate & $P(\textrm{probe}@0)$ \\
\midrule
Random                 & $3.51 \pm 0.24$ & $0.095 \pm 0.021$ & $3.50 \pm 0.14$ & $0.425 \pm 0.035$ & $0.460 \pm 0.035$ \\
\midrule
Value-greedy           & $1.29 \pm 0.02$ & $0.000 \pm 0.000$ & --- & --- & --- \\
Threshold              & $1.29 \pm 0.02$ & $0.000 \pm 0.000$ & --- & --- & --- \\
Population-myopic      & $1.90 \pm 0.27$ & $0.200 \pm 0.028$ & --- & --- & --- \\
\midrule
Personalised-myopic    & $2.84 \pm 0.28$ & $0.215 \pm 0.029$ & $4.53 \pm 0.04$ & $0.550 \pm 0.035$ & $0.000 \pm 0.000$ \\
Belief-frozen depth-2         & --- & --- & $4.54 \pm 0.04$ & $0.555 \pm 0.035$ & $0.000 \pm 0.000$ \\
\textsc{ProSE-Plan}    & $\mathbf{5.62} \pm 0.29$ & $\mathbf{0.515} \pm 0.035$ & $\mathbf{4.73} \pm 0.06$ & $\mathbf{0.815} \pm 0.028$ & $\mathbf{1.000} \pm 0.000$ \\
\midrule
Oracle depth-2         & $9.41 \pm 0.07$ & $0.990 \pm 0.007$ & $4.98 \pm 0.03$ & $0.995 \pm 0.005$ & $0.000 \pm 0.000$ \\
\bottomrule
\end{tabular}
\end{table}
\subsection{Ablation: Response-dependent Lookahead}\label{sec:exp_probe_commit}
We construct a two-step probe-commit task (Figure~\ref{fig:env_illustration}b) to isolate whether the advantage of \textsc{ProSE-Plan} comes from lookahead alone or from conditioning the second proposal on the belief induced by a possible first response.
The state space contains a start node $s_0$, two probe nodes $\{p_1,p_2\}$, and two goal nodes $\{g_1,g_2\}$, where $p_k$ lies on the path from $s_0$ to $g_k$.
The user has a hidden preference $\phi\in\{1,2\}$, with preferred goal $g_\phi$.
As in the corridor task, acceptance moves the state to the proposal, and rejection leaves the state unchanged, and we set the horizon to $T=2$.
We design the user value function such that, for the preferred probe and goal node, the value is $w_{p,+}$ and $w_{g,+}$ respectively, otherwise, the value is $w_{p,-}$ and $w_{g,-}$ respectively.
By default we use $(w_{p,+},w_{p,-},w_{g,+},w_{g,-})=(1,-3,5,4)$.
Achieving a higher success rate in this design essentially requires the assistant to learn about $\phi$:
goal proposals have high value under both preferences, so a direct commit is immediately attractive but not informative of $\phi$, while probe proposals have low immediate value, but their responses are more informative because a probe on the preferred branch is likely to be accepted, while a probe on the wrong branch is not.
Thus, probing can improve terminal success only if the planner anticipates that the first response will update the posterior and then uses that posterior to choose the second proposal.
A myopic planner has no incentive to make this sacrifice, and a belief-frozen depth-2 planner has the same horizon as \textsc{ProSE-Plan} but evaluates second-step continuations under the current belief rather than the posterior induced by hypothetical first-step responses.
This makes the task a direct test of response-dependent lookahead.

\textbf{Main results (\textbf{Q2}).}
Table~\ref{tab:final_value} shows that \textsc{ProSE-Plan} reaches $0.815$ success, compared with $0.550$ for \textsc{Personalised-myopic} and $0.555$ for \textsc{Belief-frozen depth-2}.
The behavioural signature is sharper: \textsc{ProSE-Plan} probes in every episode, while both baselines directly commit to high-value goals.
The probe response updates the posterior over $\phi$, making the second proposal more targeted and improving success by 26 percentage points.
Thus, the mechanism is not lookahead alone, but planning over how user responses update later proposal decisions.

\section{Discussion and Conclusion} \label{sec:discussion}
This work identifies \emph{evaluability} as a relevant dimension of proposal-based AI assistance: a good proposal not only has high value under the user's preferences but is also evaluable, and rejection may not mean the proposal is poor but can be information about the user's evaluation capabilities.
ProSE formalises this by treating each proposal as both a solution and an observation about the user.
Our bounded-rational response model makes this coupling explicit, and analysis of its acceptance frontier shows a concrete consequence: proposals that are likely to be accepted and those informative about evaluability need not coincide.
In particular, the most informative proposals can lie on the expected-rejection side of the acceptance frontier.
Our controlled experiments also empirically support our central claim that user evaluability is complementary to preference inference and proposal quality.
Alongside these contributions, we observe notable avenues for future work:

First, our main instantiation uses a deliberately minimal \textbf{binary accept/reject model}, and it isolates the proposal-evaluation mechanism and matches common interfaces such as accepting or dismissing a patch, edit, or revision.
However, ProSE itself is formally more general, and generalising to richer response spaces is important for more sophisticated applications.
This is possible in principle by generalising the sigmoid (binary) with a softmax (categorical decision), but future work must investigate the implications of this.
Second, the proposed \textbf{evaluability metric} captures the idea that larger or more distant changes are harder to inspect, but the right choice is most likely task-dependent.
In coding, for example, evaluability may depend on dependency structure or locality.
Learning task-specific evaluability metrics is a key next step.
Lastly, the \textbf{scalability} of \textsc{ProSE-Plan} is limited in that it is a minimal depth-2 Bayes-adaptive planner with exact grid inference.
This keeps the mechanism transparent and avoids confounding planning with approximate inference, but it does not directly scale to open-ended design spaces.
Real-world applications will require candidate generation, approximate Bayesian inference, and scalable tree search or amortized planning built upon our work.

\begin{ack}
This work was supported by the Research Council of Finland (Flagship programme: Finnish Center for Artificial Intelligence FCAI, Grant 359207), ELISE Networks of Excellence Centres (EU Horizon:2020 grant agreement 951847), and UKRI Turing AI World-Leading Researcher Fellowship (EP/W002973/1). 
We acknowledge the research environment provided by ELLIS Institute Finland. 
We also acknowledge the computational resources provided by the Aalto Science-IT Project from Computer Science IT and CSC–IT Center for Science, Finland.
\end{ack}

\bibliographystyle{unsrtnat}
\bibliography{references}

@article{steyvers2024three,
  title={Three challenges for AI-assisted decision-making},
  author={Steyvers, Mark and Kumar, Aakriti},
  journal={Perspectives on Psychological Science},
  volume={19},
  number={5},
  pages={722--734},
  year={2024},
  publisher={Sage Publications Sage CA: Los Angeles, CA}
}

@article{reddy2018shared,
  title={Shared autonomy via deep reinforcement learning},
  author={Reddy, Siddharth and Dragan, Anca D and Levine, Sergey},
  journal={arXiv preprint arXiv:1802.01744},
  year={2018}
}

@article{barke2023grounded,
  title={Grounded copilot: How programmers interact with code-generating models},
  author={Barke, Shraddha and James, Michael B and Polikarpova, Nadia},
  journal={Proceedings of the ACM on Programming Languages},
  volume={7},
  number={OOPSLA1},
  pages={85--111},
  year={2023},
  publisher={ACM New York, NY, USA}
}

@inproceedings{liang2024largescale,
  title={A large-scale survey on the usability of ai programming assistants: Successes and challenges},
  author={Liang, Jenny T and Yang, Chenyang and Myers, Brad A},
  booktitle={Proceedings of the 46th IEEE/ACM international conference on software engineering},
  pages={1--13},
  year={2024}
}

@article{ross2011bayesian,
  title={A Bayesian Approach for Learning and Planning in Partially Observable Markov Decision Processes.},
  author={Ross, St{\'e}phane and Pineau, Joelle and Chaib-Draa, Brahim and Kreitmann, Pierre},
  journal={Journal of Machine Learning Research},
  volume={12},
  number={5},
  year={2011}
}

@inproceedings{mozannar2024when,
  title={When to show a suggestion? integrating human feedback in ai-assisted programming},
  author={Mozannar, Hussein and Bansal, Gagan and Fourney, Adam and Horvitz, Eric},
  booktitle={Proceedings of the AAAI conference on artificial intelligence},
  volume={38},
  number={9},
  pages={10137--10144},
  year={2024}
}

@inproceedings{feng2025comparing,
author = {Feng, Xuening and Jiang, Zhaohui and Kaufmann, Timo and H\"{u}llermeier, Eyke and Weng, Paul and Zhu, Yifei},
title = {Comparing comparisons: informative and easy human feedback with distinguishability queries},
year = {2025},
publisher = {JMLR.org},
booktitle = {Proceedings of the 42nd International Conference on Machine Learning},
articleno = {643},
numpages = {17},
location = {Vancouver, Canada},
series = {ICML'25}
}

@article{biyik2020asking,
  title={Asking Easy Questions: A User-Friendly Approach to Active Reward Learning},
  author={Erdem Biyik and Malayandi Palan and Nicholas C. Landolfi and Dylan P. Losey and Dorsa Sadigh},
  journal={ArXiv},
  year={2019},
  volume={abs/1910.04365},
  url={https://api.semanticscholar.org/CorpusID:204008187}
}

@inproceedings{sadigh2017active,
  title={Active Preference-Based Learning of Reward Functions},
  author={Dorsa Sadigh and Anca D. Dragan and S. Shankar Sastry and Sanjit A. Seshia},
  booktitle={Robotics: Science and Systems},
  year={2017},
  url={https://api.semanticscholar.org/CorpusID:12226563}
}

@inproceedings{khoshvishkaie2023cooperative,
  title={Cooperative Bayesian optimization for imperfect agents},
  author={Khoshvishkaie, Ali and Mikkola, Petrus and Murena, Pierre-Alexandre and Kaski, Samuel},
  booktitle={Joint European Conference on Machine Learning and Knowledge Discovery in Databases},
  pages={475--490},
  year={2023},
  organization={Springer}
}

@ARTICLE{MacKay1992info,
  author={MacKay, David J. C.},
  journal={Neural Computation}, 
  title={Information-Based Objective Functions for Active Data Selection}, 
  year={1992},
  volume={4},
  number={4},
  pages={590-604},
  doi={10.1162/neco.1992.4.4.590}}

@article{Chaloner1995BayesianED,
  title={Bayesian Experimental Design: A Review},
  author={Kathryn Chaloner and Isabella Verdinelli},
  journal={Statistical Science},
  year={1995},
  volume={10},
  pages={273-304},
  url={https://api.semanticscholar.org/CorpusID:13676847}
}

@article{Simon1955ABM,
  title={A Behavioral Model of Rational Choice},
  author={Herbert A. Simon},
  journal={Quarterly Journal of Economics},
  year={1955},
  volume={69},
  pages={99-118},
  url={https://api.semanticscholar.org/CorpusID:18410595}
}

@article{Simon1956RationalCA,
  title={Rational choice and the structure of the environment.},
  author={Herbert A. Simon},
  journal={Psychological review},
  year={1956},
  volume={63 2},
  pages={
          129-38
        },
  url={https://api.semanticscholar.org/CorpusID:8503301}
}

@article{Zeckhauser1988Status,
  title={Status quo bias in decision making},
  author={Samuelson, William and Zeckhauser, Richard},
  journal={Journal of risk and uncertainty},
  volume={1},
  number={1},
  pages={7--59},
  year={1988},
  publisher={Springer}
}

@article{Sweller1988Cognitive,
author = {Sweller, John},
title = {Cognitive Load During Problem Solving: Effects on Learning},
journal = {Cognitive Science},
volume = {12},
number = {2},
pages = {257-285},
doi = {https://doi.org/10.1207/s15516709cog1202\_4},
url = {https://onlinelibrary.wiley.com/doi/abs/10.1207/s15516709cog1202_4},
eprint = {https://onlinelibrary.wiley.com/doi/pdf/10.1207/s15516709cog1202_4},
year = {1988}
}

@inproceedings{gori2026decision,
author = {Gori, Julien and Nioche, Aurelien and Johns, Christoph A. and Oulasvirta, Antti},
title = {A decision-theoretic representation of assistive interfaces},
year = {2026},
isbn = {9798400722783},
publisher = {Association for Computing Machinery},
address = {New York, NY, USA},
url = {https://doi.org/10.1145/3772318.3791819},
doi = {10.1145/3772318.3791819},
booktitle = {Proceedings of the 2026 CHI Conference on Human Factors in Computing Systems},
articleno = {1138},
numpages = {19},
location = {
},
series = {CHI '26}
}

@inproceedings{
emmons2025observation,
title={Observation Interference in Partially Observable Assistance Games},
author={Scott Emmons and Caspar Oesterheld and Vincent Conitzer and Stuart Russell},
booktitle={Forty-second International Conference on Machine Learning},
year={2025},
url={https://openreview.net/forum?id=rjZ2SWjwwB}
}

@inproceedings{
laidlaw2025assistancezero,
title={AssistanceZero: Scalably Solving Assistance Games},
author={Cassidy Laidlaw and Eli Bronstein and Timothy Guo and Dylan Feng and Lukas Berglund and Justin Svegliato and Stuart Russell and Anca Dragan},
booktitle={Forty-second International Conference on Machine Learning},
year={2025},
url={https://openreview.net/forum?id=b9hVMJi0t2}
}

@inproceedings{lou2025aaar,
author = {Lou, Renze and Xu, Hanzi and Wang, Sijia and Du, Jiangshu and Kamoi, Ryo and Lu, Xiaoxin and Xie, Jian and Sun, Yuxuan and Zhang, Yusen and Ahn, Jihyun Janice and Fang, Hongchao and Zou, Zhuoyang and Ma, Wenchao and Li, Xi and Zhang, Kai and Xia, Congying and Huang, Lifu and Yin, Wenpeng},
title = {AAAR-1.0: assessing AI's potential to assist research},
year = {2025},
publisher = {JMLR.org},
booktitle = {Proceedings of the 42nd International Conference on Machine Learning},
articleno = {1623},
numpages = {23},
location = {Vancouver, Canada},
series = {ICML'25}
}

@article{zhu2026more, 
title={More than Irrational: Modeling Belief-Biased Agents}, 
volume={40}, 
url={https://ojs.aaai.org/index.php/AAAI/article/view/40242}, 
DOI={10.1609/aaai.v40i35.40242}, 
abstractNote={Despite the explosive growth of AI and the technologies built upon it, predicting and inferring the sub-optimal behavior of users or human collaborators remains a critical challenge. In many cases, such behaviors are not a result of irrationality, but rather a rational decision made given inherent cognitive bounds and biased beliefs about the world. In this paper, we formally introduce a class of computational-rational (CR) user models for cognitively-bounded agents acting optimally under biased beliefs. The key novelty lies in explicitly modeling how a bounded memory process leads to a dynamically inconsistent and biased belief state and, consequently, sub-optimal sequential decision-making. We address the challenge of identifying the latent user-specific bound and inferring biased belief states from passive observations on the fly. We argue that for our formalized CR model family with an explicit and parameterized cognitive process, this challenge is tractable. To support our claim, we propose an efficient online inference method based on nested particle filtering that simultaneously tracks the user’s latent belief state and estimates the unknown cognitive bound from a stream of observed actions. We validate our approach in a representative navigation task using memory decay as an example of a cognitive bound. With simulations, we show that (1) our CR model generates intuitively plausible behaviors corresponding to different levels of memory capacity, and (2) our inference method accurately and efficiently recovers the ground-truth cognitive bounds from limited observations (less than 100 steps). We further demonstrate how this approach provides a principled foundation for developing adaptive AI assistants, enabling adaptive assistance that accounts for the user’s memory limitations.}, number={35}, journal={Proceedings of the AAAI Conference on Artificial Intelligence}, author={Zhu, Yifan and Katt, Sammie and Kaski, Samuel}, year={2026}, month={Mar.}, pages={29948-29956} 
}

@inproceedings{chehade2025bounded,
author = {Chehade, Mohamad and Ghosal, Soumya Suvra and Chakraborty, Souradip and Reddy, Avinash and Manocha, Dinesh and Zhu, Hao and Bedi, Amrit Singh},
title = {Bounded rationality for LLMs: satisficing alignment at inference-time},
year = {2025},
publisher = {JMLR.org},
booktitle = {Proceedings of the 42nd International Conference on Machine Learning},
articleno = {283},
numpages = {17},
location = {Vancouver, Canada},
series = {ICML'25}
}

@article{hadfield2016cooperative,
  title={Cooperative inverse reinforcement learning},
  author={Hadfield-Menell, Dylan and Russell, Stuart J and Abbeel, Pieter and Dragan, Anca},
  journal={Advances in neural information processing systems},
  volume={29},
  year={2016}
}

@article{callaway2023optimal,
  title={Optimal nudging for cognitively bounded agents: A framework for modeling, predicting, and controlling the effects of choice architectures.},
  author={Callaway, Frederick and Hardy, Mathew and Griffiths, Thomas L},
  journal={Psychological Review},
  volume={130},
  number={6},
  pages={1457},
  year={2023},
  publisher={American Psychological Association}
}

@inproceedings{depeuter2023zero,
  title={Zero-shot assistance in sequential decision problems},
  author={De Peuter, Sebastiaan and Kaski, Samuel},
  booktitle={Proceedings of the AAAI Conference on Artificial Intelligence},
  volume={37},
  pages={11551--11559},
  year={2023}
}

@inproceedings{ccelikok2022best,
author = {\c{C}elikok, Mustafa Mert and Oliehoek, Frans A. and Kaski, Samuel},
title = {Best-Response Bayesian Reinforcement Learning with Bayes-adaptive POMDPs for Centaurs},
year = {2022},
isbn = {9781450392136},
publisher = {International Foundation for Autonomous Agents and Multiagent Systems},
address = {Richland, SC},
booktitle = {Proceedings of the 21st International Conference on Autonomous Agents and Multiagent Systems},
pages = {235–243},
numpages = {9},
location = {Virtual Event, New Zealand},
series = {AAMAS '22}
}

@article{christiano2017deep,
  title={Deep reinforcement learning from human preferences},
  author={Christiano, Paul F and Leike, Jan and Brown, Tom and Martic, Miljan and Legg, Shane and Amodei, Dario},
  journal={Advances in neural information processing systems},
  volume={30},
  year={2017}
}

@article{alanqary2021modeling,
  title={Modeling the mistakes of boundedly rational agents within a Bayesian theory of mind},
  author={Alanqary, Arwa and Lin, Gloria Z and Le, Joie and Zhi-Xuan, Tan and Mansinghka, Vikash K and Tenenbaum, Joshua B},
  journal={arXiv preprint arXiv:2106.13249},
  year={2021}
}

@article{kaelbling1998planning,
  title={Planning and acting in partially observable stochastic domains},
  author={Kaelbling, Leslie Pack and Littman, Michael L and Cassandra, Anthony R},
  journal={Artificial intelligence},
  volume={101},
  number={1-2},
  pages={99--134},
  year={1998},
  publisher={Elsevier}
}

@book{duff2002optimal,
  title={Optimal Learning: Computational procedures for Bayes-adaptive Markov decision processes},
  author={Duff, Michael O'Gordon},
  year={2002},
  publisher={University of Massachusetts Amherst}
}

@article{silver2010monte,
  title={Monte-Carlo planning in large POMDPs},
  author={Silver, David and Veness, Joel},
  journal={Advances in neural information processing systems},
  volume={23},
  year={2010}
}

@article{ghavamzadeh2015bayesian,
  title={Bayesian reinforcement learning: A survey},
  author={Ghavamzadeh, Mohammad and Mannor, Shie and Pineau, Joelle and Tamar, Aviv},
  journal={Foundations and Trends{\textregistered} in Machine Learning},
  volume={8},
  number={5-6},
  pages={359--483},
  year={2015},
  publisher={Emerald Publishing Limited}
}

@inproceedings{alami2025human,
  title={Human and machine: How software engineers perceive and engage with ai-assisted code reviews compared to their peers},
  author={Alami, Adam and Ernst, Neil},
  booktitle={2025 IEEE/ACM 18th International Conference on Cooperative and Human Aspects of Software Engineering (CHASE)},
  pages={63--74},
  year={2025},
  organization={IEEE}
}

@article{ortega2013thermodynamics,
  title={Thermodynamics as a theory of decision-making with information-processing costs},
  author={Ortega, Pedro A and Braun, Daniel A},
  journal={Proceedings of the Royal Society A: Mathematical, Physical and Engineering Sciences},
  volume={469},
  number={2153},
  year={2013},
  publisher={The Royal Society}
}

@article{ortega2015information,
  title={Information-theoretic bounded rationality},
  author={Ortega, Pedro A and Braun, Daniel A and Dyer, Justin and Kim, Kee-Eung and Tishby, Naftali},
  journal={arXiv preprint arXiv:1512.06789},
  year={2015}
}

@article{fern2014decision,
  title={A decision-theoretic model of assistance},
  author={Fern, Alan and Natarajan, Sriraam and Judah, Kshitij and Tadepalli, Prasad},
  journal={Journal of Artificial Intelligence Research},
  volume={50},
  pages={71--104},
  year={2014}
}

@inproceedings{evans2016learning,
  title={Learning the preferences of ignorant, inconsistent agents},
  author={Evans, Owain and Stuhlm{\"u}ller, Andreas and Goodman, Noah},
  booktitle={Proceedings of the AAAI Conference on Artificial Intelligence},
  volume={30},
  year={2016}
}

@article{de2024preference,
  title={Preference learning of latent decision utilities with a human-like model of preferential choice},
  author={De Peuter, Sebastiaan and Zhu, Shibei and Guo, Yujia and Howes, Andrew and Kaski, Samuel},
  journal={Advances in Neural Information Processing Systems},
  volume={37},
  pages={123608--123636},
  year={2024}
}

@article{howes2023towards,
  title={Towards machines that understand people},
  author={Howes, Andrew and Jokinen, Jussi PP and Oulasvirta, Antti},
  journal={AI Magazine},
  volume={44},
  number={3},
  pages={312--327},
  year={2023},
  publisher={Wiley Online Library}
}

@inproceedings{oulasvirta2022computational,
  title={Computational rationality as a theory of interaction},
  author={Oulasvirta, Antti and Jokinen, Jussi PP and Howes, Andrew},
  booktitle={Proceedings of the 2022 CHI Conference on Human Factors in Computing Systems},
  pages={1--14},
  year={2022}
}

@article{fakhoury2024llm,
  title={Llm-based test-driven interactive code generation: User study and empirical evaluation},
  author={Fakhoury, Sarah and Naik, Aaditya and Sakkas, Georgios and Chakraborty, Saikat and Lahiri, Shuvendu K},
  journal={IEEE Transactions on Software Engineering},
  volume={50},
  number={9},
  pages={2254--2268},
  year={2024},
  publisher={IEEE}
}

@book{Sutton2018,
author = {Sutton, Richard S. and Barto, Andrew G.},
title = {Reinforcement Learning: An Introduction},
year = {2018},
isbn = {0262039249},
publisher = {A Bradford Book},
address = {Cambridge, MA, USA}
}

@article{zhi2020online,
  title={Online bayesian goal inference for boundedly rational planning agents},
  author={Zhi-Xuan, Tan and Mann, Jordyn and Silver, Tom and Tenenbaum, Josh and Mansinghka, Vikash},
  journal={Advances in neural information processing systems},
  volume={33},
  pages={19238--19250},
  year={2020}
}

@article{lieder2020resource,
  title={Resource-rational analysis: Understanding human cognition as the optimal use of limited computational resources},
  author={Lieder, Falk and Griffiths, Thomas L},
  journal={Behavioral and brain sciences},
  volume={43},
  pages={e1},
  year={2020},
  publisher={Cambridge University Press}
}

@article{lewis2014computational,
  title={Computational rationality: Linking mechanism and behavior through bounded utility maximization},
  author={Lewis, Richard L and Howes, Andrew and Singh, Satinder},
  journal={Topics in cognitive science},
  volume={6},
  number={2},
  pages={279--311},
  year={2014},
  publisher={Wiley Online Library}
}

@inproceedings{jacob2024modeling,
title={Modeling Boundedly Rational Agents with Latent Inference Budgets},
author={Athul Paul Jacob and Abhishek Gupta and Jacob Andreas},
booktitle={The Twelfth International Conference on Learning Representations},
year={2024},
url={https://openreview.net/forum?id=W3VsHuga3j}
}

@article{gershman2015computational,
  title={Computational rationality: A converging paradigm for intelligence in brains, minds, and machines},
  author={Gershman, Samuel J and Horvitz, Eric J and Tenenbaum, Joshua B},
  journal={Science},
  volume={349},
  number={6245},
  pages={273--278},
  year={2015},
  publisher={American Association for the Advancement of Science}
}

@article{li2023eliciting,
  title={Eliciting human preferences with language models},
  author={Li, Belinda Z and Tamkin, Alex and Goodman, Noah and Andreas, Jacob},
  journal={arXiv preprint arXiv:2310.11589},
  year={2023}
}

@article{ccelikok2023modeling,
  title={Modeling needs user modeling},
  author={{\c{C}}elikok, Mustafa Mert and Murena, Pierre-Alexandre and Kaski, Samuel},
  journal={Frontiers in Artificial Intelligence},
  volume={6},
  pages={1097891},
  year={2023},
  publisher={Frontiers Media SA}
}

\appendix

\section{Related Work}\label{sec:related_work}
In this section, we discuss the positioning, novelty, as well as significance of our work with related work from sequential AI assistance and assistance games, bounded and computational rationality, learning preferences and rewards from human feedback, and empirical studies of AI-assisted decision-making.
Table~\ref{tab:related_work_comparison} summarises how representative methods relate to our method.
The main takeaway from our work is that the user's response to a proposal depends on what the user can reliably evaluate from the current state, and that this evaluability is itself a latent quantity the assistant should infer and plan with.

\begin{table}[ht]
  \centering
  \caption{Where our work sits relative to representative prior work.
  \emph{Seq.} is short for sequential assistance;
  \emph{Control} means the human remains in control of the task;
  \emph{Pref.} is short for latent preference or reward inference;
  \emph{Bounded} is for bounded-rational or cognitively constrained user model;
  \emph{Eval.\ cost} is short for the response or feedback model includes a proposal evaluation cost;
  \emph{Info plan} represents planning uses the information value of the user response;
  \emph{Eval.-aware} is for proposal selection itself accounts for evaluability.
  A dash indicates Not Applicable.}
  \label{tab:related_work_comparison}
  \small
  \setlength{\tabcolsep}{4pt}
  \begin{tabular}{lccccccc}
  \toprule
  Method & Seq. & Control & Pref. & Bounded & Eval.\ cost & Info plan & Eval.-aware \\
  \midrule
  Decision-theoretic assistance \citep{fern2014decision}        & \cmark & \xmark & \cmark & \xmark & \xmark & \xmark & \xmark \\
  CIRL \citep{hadfield2016cooperative}                          & \cmark & \xmark & \cmark & \xmark & \xmark & \xmark & \xmark \\
  Shared autonomy \citep{reddy2018shared}                       & \cmark & \cmark & \cmark & \xmark & \xmark & \xmark & \xmark \\
  Centaur \citep{ccelikok2022best}                              & \cmark & \cmark & \cmark & \cmark & \xmark & \cmark & \xmark \\
  Zero-shot assistance \citep{depeuter2023zero}                 & \cmark & \cmark & \cmark & \cmark & \xmark & \cmark & \xmark \\
  AssistanceZero \citep{laidlaw2025assistancezero}              & \cmark & \xmark & \cmark & \xmark & \xmark & \xmark & \xmark \\
  Observation interference \citep{emmons2025observation}        & \cmark & \xmark & \cmark & \xmark & \xmark & \cmark & \xmark \\
  Cooperative BO \citep{khoshvishkaie2023cooperative}           & \cmark & \cmark & \cmark & \cmark & \xmark & \cmark & \xmark \\
  Optimal nudging \citep{callaway2023optimal}                   & \xmark & \cmark & \xmark & \cmark & \cmark & \xmark & \cmark \\
  RLHF \citep{christiano2017deep}                               & \xmark & ---    & \cmark & \xmark & \xmark & \xmark & \xmark \\
  Easy queries \citep{biyik2020asking}                          & \cmark & ---    & \cmark & \xmark & \cmark & \cmark & \cmark \\
  Distinguishability queries \citep{feng2025comparing}          & \cmark & ---    & \cmark & \cmark & \cmark & \cmark & \cmark \\
  When-to-show \citep{mozannar2024when}                         & \cmark & \cmark & \xmark & \xmark & \cmark & \xmark & \cmark \\
  \midrule
  \textbf{Our work}                                    & \cmark & \cmark & \cmark & \cmark & \cmark & \cmark & \cmark \\
  \bottomrule
  \end{tabular}
\end{table}

\paragraph{Sequential AI assistance and assistance games.}
A standard line of work treats AI assistance as a multi-agent decision process with hidden user goals or preferences.
Decision-theoretic assistance \citep{fern2014decision} and cooperative inverse reinforcement learning \citep{hadfield2016cooperative} establish the basic problem of acting under uncertainty about the user's reward.
Subsequent work extends the setting to shared autonomy with deep reinforcement learning \citep{reddy2018shared}, advisory assistance for bounded-rational users \citep{ccelikok2022best, depeuter2023zero}, large-scale assistance games solved with deep RL \citep{laidlaw2025assistancezero}, partially observable assistance games where assistant actions interfere with what is learned about the human \citep{emmons2025observation}, and cooperative Bayesian optimisation with imperfect agents \citep{khoshvishkaie2023cooperative}.
\citet{gori2026decision} provides a unifying decision-theoretic vocabulary for assistive interfaces, and \citet{ccelikok2023modeling} argues that user modelling is itself a load-bearing modelling task.
ProSE inherits the Bayes-adaptive structure of this line and adds proposal-dependent evaluability as a latent factor that simultaneously shapes adoption and the informativeness of the user's response.
The planner therefore has to reason about both within the same belief update, which earlier formulations do not.

\paragraph{Bounded and computational rationality.}
Bounded rationality \citep{Simon1955ABM, Simon1956RationalCA} models human behaviour as approximate optimisation under cognitive constraints.
The computational and resource-rational theoretical frameworks refine this view by deriving behaviour from rational use of limited resources \citep{lewis2014computational, gershman2015computational, lieder2020resource, oulasvirta2022computational, howes2023towards}.
Information-theoretic bounded rationality \citep{ortega2013thermodynamics, ortega2015information} formalises bounded choice as a KL-regularised problem and provides the response-model backbone we instantiate in Section~\ref{sec:response_model}.
Other work targets specific bounds, such as memory \citep{zhu2026more}, planning depth \citep{zhi2020online, alanqary2021modeling, jacob2024modeling}, or one-shot choice architectures for bounded agents \citep{callaway2023optimal}.
These models are mostly used for behavioural prediction, generation, or for one-shot intervention design as user behaviour likelihood.
We embed the same family of models inside a sequential assistance loop, where the assistant's proposal is the intervention and the user's bounded response is the observation that drives belief updates.
This turns evaluation cost from a descriptive concept into a planning quantity.

\paragraph{Learning preferences and rewards from human feedback.}
Reward learning, RLHF, and active preference learning aim to recover the user's latent objective from feedback \citep{hadfield2016cooperative, christiano2017deep, evans2016learning, sadigh2017active, biyik2020asking, li2023eliciting, de2024preference}.
\citet{evans2016learning} consider ignorant or inconsistent agents, \citet{biyik2020asking} and \citet{feng2025comparing} explicitly select queries that are easier or more distinguishable for the user, and \citet{de2024preference} models preferential choice with a human-like computational rationality choice model.
These methods focus on the inference goal and primarily treat human feedback as a query channel that does not have an impact on the transition dynamics of the task.
In ProSE, the proposal is both a candidate next state of the task and a probe of the user's latent parameters, and its evaluability shapes the response distribution.
This coupling of inference and control is what differentiates evaluability-aware proposal planning from active reward learning.

\paragraph{Empirical evidence on user evaluation cost.}
Empirical work on AI-assisted decision-making and AI-assisted programming consistently reports that users struggle with proposals that are too large or too hard to verify \citep{steyvers2024three, fakhoury2024llm, alami2025human, barke2023grounded, liang2024largescale}.
\citet{mozannar2024when} model when an AI programming assistant should show a suggestion, treating the display itself as a decision under uncertainty about user reaction.
Benchmark work on AI for research tasks finds that novel experiment proposals from LLMs often lack feasibility, making it difficult for researchers to evaluate their necessity~\citep{lou2025aaar}.
These findings motivate the central modelling choice in this paper, that proposal complexity acts as an evaluation cost in the user's response, and that an assistant ignoring this tends to produce proposals the user cannot meaningfully act upon.

\paragraph{Bayes-adaptive planning under hidden user parameters.}
Methodologically, \textsc{ProSE-Plan} is a Bayes-adaptive planner \citep{duff2002optimal, ghavamzadeh2015bayesian} for an assistance setting with a non-trivial response likelihood.
Bayes-adaptive MDPs and partially observable MDPs are standard frameworks for online inference and planning under unknown dynamics \citep{kaelbling1998planning, ross2011bayesian, silver2010monte}, and depth-limited Bayes-adaptive lookahead has been used in human-AI assistance for tractability \citep{ccelikok2022best, depeuter2023zero}.
Our depth-2 planner stays in this style but uses the response model in two roles within the same lookahead.
First, the response model scores each candidate proposal under the current belief.
Second, it updates the belief hypothetically over the user's latent parameters before scoring the follow-up proposal, so that proposals are evaluated on how their possible responses would change subsequent decisions.
This is what makes the planner sensitive to evaluability rather than only to expected acceptance.

\section{Details of Evaluability-Aware Response Model}
\subsection{Derivation of the Gibbs policy}
\label{sec:gibbs_derivation}

We derive the closed-form solution of the information-theoretic
bounded-rational decision problem in Eq.~\eqref{eq:information-theoretic-bounded-rationality}.
Let $\mathcal A$ be a finite action set, let $q\in\Delta(\mathcal A)$ be a
full-support prior policy, and let $U:\mathcal A\to\mathbb R$ be a utility
function. For $\kappa>0$, define
\begin{equation}
J(p)=\sum_{a\in\mathcal A} p(a)U(a)-\frac{1}{\kappa}\sum_{a\in\mathcal{A}}p(a)\log\frac{p(a)}{q(a)} .
\end{equation}
Since $q(a)>0$ for all $a\in\mathcal A$, the KL term is finite for all
$p\in\Delta(\mathcal A)$. Moreover,
$D_{\mathrm{KL}}(p\|q)$ is strictly convex in $p$, so $J(p)$ is strictly
concave on the simplex for $\kappa>0$. Therefore, the maximiser is unique.

To solve for it, introduce a Lagrange multiplier $\lambda$ for the simplex
constraint $\sum_{a\in\mathcal A}p(a)=1$:
\begin{equation}
\mathcal L(p,\lambda)=\sum_{a\in\mathcal{A}} p(a)U(a)-\frac{1}{\kappa}\sum_{a\in\mathcal{A}}p(a)\log\frac{p(a)}{q(a)}-\lambda(\sum_{a\in\mathcal{A}}p(a)-1).
\end{equation}
For each $a\in\mathcal A$, the first-order condition is
\begin{equation}
\frac{\partial \mathcal{L}}{\partial p(a)}=U(a)-\frac{1}{\kappa}(\log\frac{p(a)}{q(a)}+1)-\lambda=0.
\end{equation}
Rearranging gives
\begin{equation}
\log\frac{p(a)}{q(a)}=\kappa U(a)-\kappa\lambda-1 .
\end{equation}
Exponentiating both sides,
\begin{equation}
p(a)=q(a)\exp(\kappa U(a)-\kappa\lambda-1)=q(a)\exp(\kappa U(a))\exp(-\kappa\lambda-1).
\end{equation}
The factor $\exp(-\kappa\lambda-1)$ does not depend on $a$, so it is fixed by normalisation. 
Enforcing $\sum_{a\in\mathcal A}p(a)=1$ yields
\begin{equation}
Z_\kappa =\sum_{a'\in\mathcal{A}}q(a')\exp(\kappa U(a')),
\end{equation}
and therefore
\begin{equation}
p^{\mathrm{ITBR}}(a)=
\frac{q(a)\exp(\kappa U(a))}{Z_\kappa}.
\end{equation}
This is the Gibbs policy used in Eq.~\eqref{eq:itbr-gibbs}. It can be read as the prior policy $q$ tilted toward higher-utility actions by the factor $\exp(\kappa U(a))$.

Finally, the default-policy limit follows directly from the same expression:
\begin{equation}
\lim_{\kappa\to 0^+}
p^{\mathrm{ITBR}}(a)= \frac{q(a)}{\sum_{a'\in\mathcal A}q(a')} = q(a).
\end{equation}
Thus, when the information-processing parameter vanishes, the agent does not move away from the prior policy.

\subsection{Reduction to a binary sigmoid response}
\label{app:binary-logistic}
We now specialise the general Gibbs policy to the binary response space used in this work
$\mathcal Y=\{\mathrm{acc},\mathrm{rej}\}$.
Suppressing the conditioning variables for readability, the acceptance probability is
\begin{equation}
    p^\star(\mathrm{acc})
    =
    \frac{
        q(\mathrm{acc})\exp(\kappa U(\mathrm{acc}))
    }{
        q(\mathrm{acc})\exp(\kappa U(\mathrm{acc}))
        +
        q(\mathrm{rej})\exp(\kappa U(\mathrm{rej}))
    }.
    \label{eq:app-binary-gibbs}
\end{equation}
Dividing the Gibbs weights for acceptance and rejection yields
\begin{equation}
    \frac{
        p^\star(\mathrm{acc})
    }{
        p^\star(\mathrm{rej})
    }
    =
    \frac{
        q(\mathrm{acc})
    }{
        q(\mathrm{rej})
    }
    \exp
    (
        \kappa[
            U(\mathrm{acc})-U(\mathrm{rej})
        ]
    ).
    \label{eq:app-binary-odds}
\end{equation}
Let
$\Delta U=U(\mathrm{acc})-U(\mathrm{rej})$
and
$\lambda=\log\frac{q(\mathrm{acc})}{q(\mathrm{rej})}$, we thus obtain Eq.~\eqref{eq:binary-response-sigmoid}:
\begin{equation}
    p^\star(\mathrm{acc})
    =
    \sigma(
        \kappa\Delta U+\lambda
    ).
    \label{eq:app-binary-sigmoid}
\end{equation}

\subsection{Response utilities}
\label{app:response-utilities}

Fix a current state $s$, proposal $\tilde{s}$, and preference parameter $\phi$.
For a general response $y\in\mathcal Y$, define response utility as the expected value of the realised next state:
\begin{equation}
    U(y;s,\tilde{s},\phi)
    =
    \mathbb E_{s'\sim T^y(\cdot\mid s,\tilde{s},y)}
    [
        V_\phi(s')
    ].
    \label{eq:app-response-utility}
\end{equation}
In the binary deterministic instantiation,
\begin{equation}
    T^y(s'\mid s,\tilde{s},\mathrm{acc})
    =
    \mathbf{1}\{s'=\tilde{s}\},
    \qquad
    T^y(s'\mid s,\tilde{s},\mathrm{rej})
    =
    \mathbf{1}\{s'=s\}.
    \label{eq:app-binary-transition}
\end{equation}
Therefore
\begin{equation}
    U(\mathrm{acc};s,\tilde{s},\phi)
    =
    V_\phi(\tilde{s}),
    \qquad
    U(\mathrm{rej};s,\tilde{s},\phi)
    =
    V_\phi(s),
    \label{eq:app-binary-utilities}
\end{equation}
and
\begin{equation}
    \Delta U_\phi(s,\tilde{s})
    =
    V_\phi(\tilde{s})-V_\phi(s)
    =
    \Delta V_\phi(s,\tilde{s}).
    \label{eq:app-value-gain}
\end{equation}

\subsection{Distance-dependent default response}
\label{app:default-response}

The binary default policy is fully determined by its log-odds
\begin{equation}
    \lambda_\theta(s,\tilde{s})
    =
    \log
    \frac{
        q_\theta(\mathrm{acc}\mid s,\tilde{s})
    }{
        q_\theta(\mathrm{rej}\mid s,\tilde{s})
    }.
    \label{eq:app-default-logodds}
\end{equation}
To encode proposal-dependent evaluability, we use a monotone distance penalty:
\begin{equation}
    \lambda_\theta(s,\tilde{s})
    =
    -
    \alpha g(d(s,\tilde{s})),
    \label{eq:app-distance-logodds-general}
\end{equation}
where $d(s,\tilde{s})\ge0$ is proposal distance, and $g$ is a nondecreasing burden transform with $g(0)=0$.

\section{Proofs and Details for the Information Frontier}
\label{app:information-frontier-proof}

\subsection{Information frontier conditional on user parameters}
\label{app:conditional-information-frontier}

We prove Proposition~\ref{prop:conditional_information_frontier}.
The result is conditional on a latent user-parameter $z=(\phi,\rho,\kappa)$ with $\rho>0$ and $\kappa>0$.
This does not assume that the assistant knows $z$, but analyses the response likelihood pointwise in $z$; the assistant later integrates the same likelihood under its posterior belief.

Consider a proposal path $\{\tilde{s}(t)\}_{t\ge0}$ from current state $s$, parameterised by distance so that $\tilde{s}(0)=s$ and $d(s,\tilde{s}(t))=t$.
Along this path, define
\begin{equation}
    \eta_z(t) = \kappa[\Delta V_\phi(t)-\rho g(t)],
    \qquad
    \Delta V_\phi(t) := V_\phi(\tilde{s}(t))-V_\phi(s),
    \label{eq:app-path-logit}
\end{equation}
and let
\begin{equation}
    p_z(t) = \sigma(\eta_z(t))
    \label{eq:app-path-acceptance-probability}
\end{equation}
be the acceptance probability at distance $t$.

\paragraph{Deriving the Fisher information.}
For fixed $t$ and $z$, the binary response can be written as
$Y_t\in\{0,1\}$, where $Y_t=1$ denotes acceptance and
$Y_t=0$ denotes rejection.
Thus
\begin{equation}
    Y_t \sim \mathrm{Bernoulli}(p_z(t)).
\end{equation}
To derive the Fisher information about $\rho$, suppress the dependence on $t$ and $z$ for readability and write $p=p_z(t)$.
The Bernoulli likelihood is
\begin{equation}
    L(\rho;Y) = p^Y(1-p)^{1-Y},
\end{equation}
and the log-likelihood is
\begin{equation}
    \ell(\rho;Y) = Y\log p + (1-Y)\log(1-p).
\end{equation}
Differentiating with respect to $\rho$ gives the score
\begin{align}
    \frac{\partial \ell}{\partial \rho}
    &=
    Y\frac{1}{p}\frac{\partial p}{\partial \rho}
    -
    (1-Y)\frac{1}{1-p}\frac{\partial p}{\partial \rho}
    \\
    &=
    \frac{Y-p}{p(1-p)}
    \frac{\partial p}{\partial \rho}.
    \label{eq:app-bernoulli-score}
\end{align}
The Fisher information is the expected squared score:
\begin{equation}
    I^\rho_z(t) = \mathbb E_Y[(\frac{\partial \ell}{\partial\rho})^2].
\end{equation}
Using Eq.~\eqref{eq:app-bernoulli-score},
\begin{align}
    I^\rho_z(t)
    &=
    \mathbb E_Y
    [
        \frac{(Y-p)^2}{p^2(1-p)^2}
    ]
    \left(
        \frac{\partial p}{\partial \rho}
    \right)^2
    \\
    &=
    \frac{
        \mathrm{Var}(Y)
    }{
        p^2(1-p)^2
    }
    \left(
        \frac{\partial p}{\partial \rho}
    \right)^2
    \\
    &=
    \frac{
        \left(
            \frac{\partial p}{\partial \rho}
        \right)^2
    }{
        p(1-p)
    },
    \label{eq:app-bernoulli-fisher-general}
\end{align}
because $\mathrm{Var}(Y)=p(1-p)$ for a Bernoulli random variable.

Since $p_z(t)=\sigma(\eta_z(t))$ and $\sigma'(\eta)=\sigma(\eta)(1-\sigma(\eta))$,
\begin{align}
    \frac{\partial p_z(t)}{\partial \rho}
    &=
    \sigma'(\eta_z(t))
    \frac{\partial \eta_z(t)}{\partial \rho}
    \\
    &=
    p_z(t)(1-p_z(t))
    \left[
        -\kappa g(t)
    \right].
    \label{eq:app-p-derivative-rho}
\end{align}
Substituting Eq.~\eqref{eq:app-p-derivative-rho} into
Eq.~\eqref{eq:app-bernoulli-fisher-general} gives Eq.~\eqref{eq:rho_fisher_information}:
\begin{equation}
    I^\rho_z(t)
    =
    \kappa^2 g(t)^2
    p_z(t)(1-p_z(t)).
    \label{eq:app-rho-fisher}
\end{equation}

\paragraph{Interpretation.}
Equation~\eqref{eq:app-rho-fisher} separates informativeness into two factors.
The term $g(t)^2$ is the squared sensitivity of the logit to $\rho$.
It is small near the current state because a small-distance proposal is almost insensitive to the evaluability slope.
The term $p_z(t)(1-p_z(t))$ is the Bernoulli response variance.
It is large when acceptance and rejection are both plausible and small when the response is nearly deterministic.
The information frontier arises from the product of these two terms.

\paragraph{Boundary behaviour.}
At $t=0$, $g(0)=0$, so Eq.~\eqref{eq:app-rho-fisher} gives
\begin{equation}
    I^\rho_z(0)=0.
\end{equation}

For the tail, assume $\Delta V_\phi(t)=o(g(t))$ as $t\to\infty$.
Then for any $\epsilon>0$, there exists $T$ such that
$\Delta V_\phi(t)\le \epsilon g(t)$ for all $t>T$.
Choose $\epsilon=\rho/2$.
Then for all $t>T$,
\begin{align}
    \eta_z(t)
    &=
    \kappa[\Delta V_\phi(t)-\rho g(t)]
    \\
    &\le
    -\frac{\kappa\rho}{2}g(t).
    \label{eq:app-tail-logit-bound}
\end{align}
For $\eta\le0$, we have
$\sigma(\eta)(1-\sigma(\eta))\le\sigma(\eta)\le e^\eta$.
Thus,
\begin{align}
    I^\rho_z(t)
    &=
    \kappa^2 g(t)^2
    \sigma(\eta_z(t))
    [1-\sigma(\eta_z(t))]
    \\
    &\le
    \kappa^2 g(t)^2
    \exp
    \left(
        -\frac{\kappa\rho}{2}g(t)
    \right).
    \label{eq:app-tail-information-bound}
\end{align}
Since $g(t)\to\infty$ and $x^2e^{-cx}\to0$ for any $c>0$, the right-hand side tends to zero.
Therefore
\begin{equation}
    I^\rho_z(t)\to0
    \qquad
    \text{as}
    \qquad
    t\to\infty.
\end{equation}

\paragraph{Interior maximum.}
The function $I^\rho_z(t)$ is continuous because $g$ and $\Delta V_\phi$ are continuous.
For any $t_0>0$, we have $g(t_0)>0$ and
$p_z(t_0)(1-p_z(t_0))>0$, so $I^\rho_z(t_0)>0$.
Since $I^\rho_z(t)\to0$ as $t\to\infty$, there exists $R>t_0$ such that
$I^\rho_z(t)<I^\rho_z(t_0)/2$ for all $t>R$.
By continuity, $I^\rho_z$ attains a maximum on the compact interval $[0,R]$.
Since $I^\rho_z(0)=0<I^\rho_z(t_0)$, this maximum is attained at some
$t^\star\in(0,R]$, and the tail bound makes it a global maximum over $[0,\infty)$.

\paragraph{Rejection-side characterization.}
Now we show that every global maximiser satisfies $\eta_z(t^\star)<0$, equivalently $p_z(t^\star)<1/2$ with Proof by Elimination.

First suppose $\eta_z(t^\star)>0$.
Because $\Delta V_\phi(t)=o(g(t))$ and $\rho>0$, we have
$\eta_z(t)\to-\infty$ as $t\to\infty$.
By continuity, there exists $t_F>t^\star$ such that
$\eta_z(t_F)=0$.
This $t_F$ is the acceptance-frontier distance along the path.
Since $g$ is strictly increasing, $g(t_F)>g(t^\star)$.
At $t_F$, the Bernoulli variance is maximal:
\begin{equation}
    p_z(t_F)(1-p_z(t_F))
    =
    \frac{1}{4}.
\end{equation}
At $t^\star$, since $\eta_z(t^\star)\neq0$, we have
\begin{equation}
    p_z(t^\star)(1-p_z(t^\star))
    <
    \frac{1}{4}.
\end{equation}
Therefore
\begin{align}
I^\rho_z(t_F)
&=
\kappa^2 g(t_F)^2 \frac{1}{4}
\\
&>
\kappa^2 g(t^\star)^2 \frac{1}{4}
\\
&>
\kappa^2 g(t^\star)^2
p_z(t^\star)(1-p_z(t^\star))
\\
&=
I^\rho_z(t^\star),
\end{align}
contradicting the global optimality of $t^\star$.

Now suppose $\eta_z(t^\star)=0$.
Since $t^\star>0$, we have $g(t^\star)>0$.
Let
\begin{equation}
v(\eta)
=
\sigma(\eta)(1-\sigma(\eta)).
\end{equation}
Then
\begin{equation}
I^\rho_z(t)
=
\kappa^2g(t)^2v(\eta_z(t)).
\end{equation}
Differentiating with respect to $t$ gives
\begin{equation}
\frac{d}{dt}I^\rho_z(t)
=
\kappa^2
\left[
    2g(t)g'(t)v(\eta_z(t))
    +
    g(t)^2v'(\eta_z(t))\eta'_z(t)
\right].
\label{eq:app-information-derivative}
\end{equation}
The derivative of the Bernoulli variance is
\begin{equation}
v'(\eta)
=
\sigma(\eta)(1-\sigma(\eta))(1-2\sigma(\eta)).
\end{equation}
At $\eta=0$, $\sigma(0)=1/2$, so $v'(0)=0$ and $v(0)=1/4$.
Therefore, at $t^\star$,
\begin{equation}
\frac{d}{dt}I^\rho_z(t^\star)
=
\kappa^2
\cdot
2g(t^\star)g'(t^\star)
\cdot
\frac{1}{4}
>
0,
\end{equation}
where the strict inequality uses $g(t^\star)>0$ and $g'(t^\star)>0$.
Thus $t^\star$ cannot be a local maximum, again a contradiction.

Both cases lead to contradictions, so every global maximiser satisfies
$\eta_z(t^\star)<0$.
Equivalently,
\begin{equation}
p_z(t^\star)
=
\sigma(\eta_z(t^\star))
<
\frac{1}{2}.
\end{equation}
This proves Proposition~\ref{prop:conditional_information_frontier}.

\subsection{Relation to mutual information and value of information}
\label{app:mi-voi}

The $\rho$-Fisher information in Proposition~\ref{prop:conditional_information_frontier} is a local analytic proxy for how proposal distance affects learning about evaluability.
A more general Bayesian measure of response informativeness is the mutual information between the latent user parameter and the response:
\begin{equation}
    I_b(z;y\mid s,\tilde{s}) = \mathbb E_y[D_{\mathrm{KL}}(b^y \| b)],
    \label{eq:app-mutual-information}
\end{equation}
where $b^y$ is the posterior after observing response $y$ to proposal $\tilde{s}$.

This quantity is closer to the Bayesian planner's actual information gain, because it measures how much the response changes the assistant's belief over the whole latent parameter vector $z$.
The Fisher information used in the main analysis is narrower: it asks how sensitive one binary response is to the evaluability slope $\rho$ at a given parameter vector.
We use Fisher information in the theorem because it gives a closed-form local diagnostic of how proposal distance affects learning.

To connect mutual information to downstream planning value, suppose the continuation value function is $M$-Lipschitz in the belief under total variation distance.
Then the downstream value change induced by the belief update satisfies
\begin{equation}
    |\mathrm{VoI}(s,\tilde{s},b)|\le M \mathbb E_y[\|b^y-b\|_{\mathrm{TV}}].
\end{equation}
By Pinsker's inequality,
\begin{equation}
    \|b^y-b\|_{\mathrm{TV}}\le \sqrt{ \frac{1}{2}D_{\mathrm{KL}}(b^y\|b)}.
\end{equation}
Applying Jensen's inequality gives
\begin{equation}
    |\mathrm{VoI}(s,\tilde{s},b)| \le M \sqrt{\frac{1}{2}I_b(z;y\mid s,\tilde{s})}.
    \label{eq:app-voi-mi-bound}
\end{equation}
Thus mutual information controls the possible downstream value of belief updates, while $\rho$-Fisher information gives a tractable local view of how proposal distance affects learning about evaluability.

\section{Planner Details}
\label{app:planner-details}

\subsection{Finite-grid belief tracking}
\label{app:finite-grid-belief}

The planner maintains a belief over latent user parameters $z=(\phi,\rho,\kappa)$.
In the main experiments, we approximate the latent parameter space by a finite grid
$\mathcal Z_{\mathrm{grid}}$, and maintain the posterior exactly on this grid.
For a current belief $b$, state $s$, proposal $\tilde{s}$, and observed response $y$, the grid posterior is
\begin{equation}
b^y_{s,\tilde{s}}(z)
=
\frac{
P(y\mid s,\tilde{s},z)b(z)
}{
\sum_{z'\in\mathcal Z_{\mathrm{grid}}}
P(y\mid s,\tilde{s},z')b(z')
}.
\label{eq:app-response-belief-update}
\end{equation}
The corresponding predictive response distribution is
\begin{equation}
p_b(y\mid s,\tilde{s})
=
\sum_{z\in\mathcal Z_{\mathrm{grid}}}
P(y\mid s,\tilde{s},z)b(z).
\label{eq:app-predictive-response}
\end{equation}
Thus the finite-grid approximation only concerns the representation of $b$; once the grid is fixed, the Bayesian update is exact on that grid.

\subsection{Response-branch and transition-branch views}
\label{app:response-transition-views}

The main text presents \textsc{ProSE-Plan} in terms of hypothetical next states $s'$, while the belief update in Eq.~\eqref{eq:prose-belief-update} is written in terms of the observed response $y$.
These are two equivalent views in the binary deterministic instantiation.

In the general ProSE process, for fixed user parameter $z$, the response model and response-mediated transition induce an assistant-side transition kernel
\begin{equation}
\mathcal T^{\mathrm{AI}}_z(s'\mid s,\tilde{s})
=
\sum_{y\in\mathcal Y}
P(y\mid s,\tilde{s},z)
\mathcal T^y(s'\mid s,\tilde{s},y).
\label{eq:app-induced-transition}
\end{equation}
Under belief $b$, the predictive next-state distribution is
\begin{equation}
p(s'\mid b,s,\tilde{s})
=
\sum_{z\in\mathcal Z_{\mathrm{grid}}}
b(z)
\mathcal T^{\mathrm{AI}}_z(s'\mid s,\tilde{s})
=
\sum_{y\in\mathcal Y}
p_b(y\mid s,\tilde{s})
\mathcal T^y(s'\mid s,\tilde{s},y).
\label{eq:app-predictive-next-state}
\end{equation}
This is the quantity used in Eq.~\eqref{eq:planner-transition-likelihood}.

If the response $y$ is observed, the posterior is Eq.~\eqref{eq:app-response-belief-update}.
If only the realised next state $s'$ is observed, the posterior can instead be written as
\begin{equation}
b^{s'}_{s,\tilde{s}}(z)
=
\frac{
\mathcal T^{\mathrm{AI}}_z(s'\mid s,\tilde{s})b(z)
}{
\sum_{z'\in\mathcal Z_{\mathrm{grid}}}
\mathcal T^{\mathrm{AI}}_{z'}(s'\mid s,\tilde{s})b(z')
}.
\label{eq:app-transition-belief-update}
\end{equation}

In the binary deterministic response model,
\begin{equation}
\mathcal Y=\{\mathrm{acc},\mathrm{rej}\},
\qquad
s^{\mathrm{acc}}=\tilde{s},
\qquad
s^{\mathrm{rej}}=s.
\label{eq:app-binary-next-state}
\end{equation}
Thus, when $\tilde{s}\neq s$, observing $s'=\tilde{s}$ is equivalent to observing $y=\mathrm{acc}$, and observing $s'=s$ is equivalent to observing $y=\mathrm{rej}$.
Therefore,
\begin{equation}
b^{\tilde{s}}_{s,\tilde{s}} = b^{\mathrm{acc}}_{s,\tilde{s}},
\qquad
b^{s}_{s,\tilde{s}} = b^{\mathrm{rej}}_{s,\tilde{s}}.
\label{eq:app-state-response-equivalence}
\end{equation}
This is why Algorithm~\ref{alg:prose-plan} can branch over the two next-state outcomes $(s,\tilde{s})$, while still implementing response-dependent belief updates.
If self-proposals $\tilde{s}=s$ are allowed, the next state alone no longer identifies the response; in that case the planner should branch directly on $y$, or the response should be treated as observed.

\subsection{Depth-2 expansion}
\label{app:depth-two-expansion}

We now spell out the depth-2 score used by \textsc{ProSE-Plan}.
Define the belief-averaged terminal value
\begin{equation}
H(s,b)
=
\sum_{z\in\mathcal Z_{\mathrm{grid}}}
b(z)V_\phi(s).
\label{eq:app-leaf-value}
\end{equation}
This is the precise meaning of the terminal condition used in the planner: at a leaf node, the planner evaluates the state by averaging the user's latent value $V_\phi(s)$ under the current belief.

For a one-step proposal $\tilde{s}$ from state $s$ under belief $b$, define
\begin{equation}
Q_1(s,b,\tilde{s})
=
\sum_{y\in\mathcal Y}
p_b(y\mid s,\tilde{s})
\sum_{s'\in\mathcal S}
\mathcal T^y(s'\mid s,\tilde{s},y)
H(s',b^y_{s,\tilde{s}}).
\label{eq:app-one-step-score-general}
\end{equation}
Under the binary deterministic transition, this simplifies to
\begin{equation}
Q_1(s,b,\tilde{s})
=
p_b(\mathrm{acc}\mid s,\tilde{s})
H(\tilde{s},b^{\mathrm{acc}}_{s,\tilde{s}})
+
p_b(\mathrm{rej}\mid s,\tilde{s})
H(s,b^{\mathrm{rej}}_{s,\tilde{s}}).
\label{eq:app-one-step-score-binary}
\end{equation}

The depth-2 score of a first proposal $\tilde{s}$ is obtained by considering each possible first response, updating the belief under that response, and choosing the best second proposal:
\begin{equation}
Q_2(s,b,\tilde{s})
=
\sum_{y\in\mathcal Y}
p_b(y\mid s,\tilde{s})
\sum_{s'\in\mathcal S}
\mathcal T^y(s'\mid s,\tilde{s},y)
\max_{\tilde{s}'\in\mathcal C(s')}
Q_1(s',b^y_{s,\tilde{s}},\tilde{s}').
\label{eq:app-depth-two-score-general}
\end{equation}
In the binary deterministic case, Eq.~\eqref{eq:app-depth-two-score-general} becomes
\begin{align}
Q_2(s,b,\tilde{s})
=&\;
p_b(\mathrm{acc}\mid s,\tilde{s})
\max_{\tilde{s}'\in\mathcal C(\tilde{s})}
Q_1(\tilde{s},b^{\mathrm{acc}}_{s,\tilde{s}},\tilde{s}')
\nonumber\\
&+
p_b(\mathrm{rej}\mid s,\tilde{s})
\max_{\tilde{s}'\in\mathcal C(s)}
Q_1(s,b^{\mathrm{rej}}_{s,\tilde{s}},\tilde{s}').
\label{eq:app-depth-two-score-binary}
\end{align}
Algorithm~\ref{alg:prose-plan} is an implementation of Eq.~\eqref{eq:app-depth-two-score-binary}, written in terms of the two possible next states $s'=\tilde{s}$ and $s'=s$.

\begin{algorithm}[t]
\caption{\textsc{ProSE-Plan}: depth-2 proposal selection}
\label{alg:prose-plan}
\begin{algorithmic}[1]
\Require State $s$, belief $b$, candidate map $\mathcal{C}$,  transition $\mathcal{T}$
\For{each candidate proposal $\tilde{s} \in \mathcal{C}(s)$}
    \For{both outcomes $s' \in (s, \tilde s)$}
        \State $p_{s'} \gets  \mathrm{E}_{z \sim b} \big[ \mathcal{T}(s' \mid s, \tilde{s}) \big]$ \Comment{Eq.\eqref{eq:planner-transition-likelihood}} \label{alg-line:user-likelihood}
        \State $b_{s'} \gets \textsc{BayesUpdate}(b;\, s, \tilde{s}, s')$ \Comment{Eq.~\eqref{eq:prose-belief-update}}
        \State $Q^{s'}_2(s, b, \tilde s) \gets \max_{\tilde{s}' \in \mathcal{C}(s')} Q^*_1(s', b_{s'}, \tilde s')$ \Comment{Eq.~\eqref{eq:bamdp-q}, where $Q_0 = V_\phi$} \label{alg-line:q2}
    \EndFor
    \State $Q_2(s, b, \tilde s) = \sum_{s'} p_{s'} \times Q^{s'}_2(s, b, \tilde s)$
\EndFor
\State \textbf{return} $\argmax_{\tilde{s} \in \mathcal{C}(s)} Q_2(\tilde{s})$
\end{algorithmic}
\end{algorithm}

\subsection{Relation to the Bayes-adaptive recursion}
\label{app:planner-bamdp-relation}

Eq.~\eqref{eq:app-depth-two-score-general} is the $h=2$ finite-lookahead instantiation of the Bayes-adaptive recursion in Eq.~\eqref{eq:bamdp-q}.
The generic action $a$ in Eq.~\eqref{eq:bamdp-q} is a proposal $\tilde{s}\in\mathcal C(s)$ in ProSE.
The generic latent model parameter is the latent user parameter $z=(\phi,\rho,\kappa)$.
The transition likelihood in the generic BA-MDP is replaced by the proposal-induced transition kernel $\mathcal T_z^{\mathrm{AI}}$ in Eq.~\eqref{eq:app-induced-transition}, or equivalently by the response likelihood $P(y\mid s,\tilde{s},z)$ when the response is observed.
The terminal value is $H(s,b)$ in Eq.~\eqref{eq:app-leaf-value}, rather than an additional immediate reward at each step, because our experiments use a terminal-artefact objective.

\subsection{Computational complexity}
\label{app:planner-complexity}

Let $|\mathcal C|$ be the maximum candidate-set size, $|\mathcal Y|$ the number of possible responses, and $|\mathcal Z_{\mathrm{grid}}|$ the number of grid points in the belief.
Computing either $p_b(y\mid s,\tilde{s})$ or $H(s,b)$ requires summing over the grid, and therefore costs $O(|\mathcal Z_{\mathrm{grid}}|)$.
For each first-step proposal, \textsc{ProSE-Plan} considers $|\mathcal Y|$ first responses; for each response, it evaluates up to $|\mathcal C|$ second proposals; and each second proposal considers $|\mathcal Y|$ responses.
Thus exhaustive depth-2 evaluation costs
\[
O(|\mathcal C|^2|\mathcal Y|^2|\mathcal Z_{\mathrm{grid}}|)
\]
per planning step under deterministic response-mediated transitions.
In our experiments, $|\mathcal Y|=2$, so the main scaling factors are the candidate-set size and the grid size.

\section{Experiment Details}
\label{app:experiments}

This appendix provides implementation details, full results, and sensitivity checks for the experiments in Section~\ref{sec:experiment}.

\begin{table}[t]
  \centering
    \caption{%
    \textbf{Baseline methods.}
    Each method removes one or more components of \textsc{ProSE-Plan}.
    }
  \label{tab:baselines}
  \small
  \begin{tabular}{lcccccc}
  \toprule
  \makecell{Method}
  & \makecell{Role}
  & \makecell{Preference}
  & \makecell{Evaluability}
  & \makecell{Personal.\\$\rho,\kappa$}
  & \makecell{Depth-2 \\ lookahead}
  & \makecell{Resp.-dep.\\lookahead update} \\
  \midrule
  Random  &baseline & none    & none         & \xmark & \xmark & \xmark  \\
  Value-greedy  &baseline & posterior    & none         & \xmark & \xmark & \xmark  \\
  Threshold~\cite{chehade2025bounded}  &baseline    & posterior    & fixed threshold & \xmark & \xmark & \xmark   \\
  Population-myopic   &baseline  & posterior    & fixed $(\bar{\rho},\bar{\kappa})$  & \xmark & \xmark & \xmark   \\
  Personalized-myopic     &baseline   & posterior     & posterior    & \cmark & \xmark & \xmark   \\
  Belief-frozen\\depth-2  &ablation  & posterior & posterior    & \cmark & \cmark & \xmark   \\
  \textsc{ProSE-Plan}&full method & posterior & posterior    & \cmark & \cmark & \cmark  \\
  Oracle depth-2  &reference & true $z$  & true $z$     & ---    & \cmark & ---      \\
  \bottomrule
  \end{tabular}
  \end{table}

\subsection{Baseline Definitions}
\label{app:baselines}
All methods operate on the same candidate set $\mathcal C(s_t)$ unless otherwise stated.
For a belief $b$, define the belief-averaged state value
\begin{equation*}
H(s,b)
=
\sum_{z\in\mathcal Z_{\mathrm{grid}}}
b(z)V_\phi(s).
\end{equation*}
For a proposal $\tilde{s}$ from state $s$, define the belief-predictive response probability
\begin{equation*}
p_b(y\mid s,\tilde{s})
=
\sum_{z\in\mathcal Z_{\mathrm{grid}}}
P(y\mid s,\tilde{s},z)b(z),
\end{equation*}
and let $b^y_{s,\tilde{s}}$ denote the posterior after hypothetically observing response $y$.
Under the binary deterministic transition, write
\begin{equation*}
s^y
=
\begin{cases}
\tilde{s}, & y=\mathrm{acc},\\
s, & y=\mathrm{rej}.
\end{cases}
\end{equation*}
\paragraph{Random.}
\begin{equation*}
  \tilde{s} \sim U(\mathcal{C}(s)).
\end{equation*}
The random baseline samples uniformly from the candidate set and does not use the value function, response model, belief state, proposal distance, or planning horizon.
It serves as a null policy for checking that the task is nontrivial and that successful performance is not an artifact of candidate-set construction or graph topology alone.
Observed responses are still generated by the same user model as in all other conditions, but the random planner does not use those responses to choose future proposals.

\paragraph{Value-greedy.}
\begin{equation*}
  \tilde{s}_* = \arg\max_{\tilde{s} \in \mathcal{C}(s)} \mathbb{E}_{\phi \sim b_t} [ V_\phi(\tilde{s}) ] = \arg\max_{\tilde{s} \in \mathcal{C}(s_t)}
  \sum_\phi b_t(\phi) V_\phi(\tilde{s}).
\end{equation*}
The value-greedy baseline selects the candidate with the highest posterior-expected user value under the current belief over the preferred branch.
%Equivalently, because the current state $s_t$ is fixed, this also maximises expected value gain over the full candidate set.
This baseline tests whether the task can be solved by preference learning and value maximisation alone, without modelling evaluability constraints.
It ignores proposal distance in the planning objective and uses a fixed response model with $\rho=0$ and $\kappa=1$; therefore, rejections are interpreted as evidence about $\phi$ rather than evidence that the proposal was too difficult to evaluate.

\paragraph{Threshold.}
\begin{equation*}
  \tilde{s}_* = \arg\max_{\tilde{s} \in \mathcal{C}_\tau(s_t)} \mathbb{E}_{\phi \sim b_t}[V_\phi(\tilde{s})] = \arg\max_{\tilde{s} \in \mathcal{C}_\tau(s_t)}
  \sum_\phi b_t(\phi) V_\phi(\tilde{s}).
\end{equation*}
where $\mathcal{C}_\tau(s_t) = \{ s \in \mathcal{C}(s_t): d(s_t, s) \le \tau \}$.
This baseline is a distance-constrained version of the Value-greedy planner.
It represents a simple fixed-locality heuristic for evaluability.
Rather than modelling or inferring the user's distance sensitivity, the assistant removes proposals that are farther than a global threshold and then applies the same value-greedy rule as Value-greedy.

The baseline tests whether the benefits of evaluability-aware planning can be explained by a hand-tuned locality constraint alone, without personalised inference over $\rho$ or response-contingent lookahead.
As with Value-greedy, it updates the belief over $\phi$ from observed responses but uses a fixed response model with $\rho=0$ and $\kappa=1$.
Because $\tau$ is a free heuristic parameter rather than a parameter learned from the response model, we tune it via grid search on held-out validation seeds.
%This gives the threshold baseline a strong, non-arbitrary cutoff while keeping the reported main results separate from model selection.
For the corridor experiments, we evaluate $\tau \in \{1,\ldots,8\}$ on validation seeds $0$--$49$, across $\rho_{\mathrm{true}}\in\{0.04,0.08,0.18,0.30,0.36\}$, $\alpha_{\mathrm{env}}\in\{0,0.25,0.5,0.75,1\}$, and $\kappa_{\mathrm{true}}=1.0$.
%The selected threshold is global across corridor conditions rather than re-tuned separately for each test setting.
We choose the threshold with the highest mean terminal value, breaking ties by success rate and then by the smaller threshold, so equally good thresholds favour the more local and conservative rule.
This procedure selects $\tau=4.0$; in the validation run, $\tau=4,\ldots,8$ tie in mean terminal value and success rate, so the smallest tied value is used.
If no candidate satisfies the cutoff, the method falls back to the nearest available candidate.

\paragraph{Population myopic.}
\begin{equation*}
  \tilde{s}_*
  =
  \arg\max_{\tilde{s} \in \mathcal{C}(s_t)}
  \sum_\phi b_t(\phi)
  \Big[
    P_{\bar{\rho},\bar{\kappa}}(\mathrm{acc}\mid s_t,\tilde{s},\phi)
    V_\phi(\tilde{s})
    +
    \big(1-P_{\bar{\rho},\bar{\kappa}}(\mathrm{acc}\mid s_t,\tilde{s},\phi)\big)
    V_\phi(s_t)
  \Big],
\end{equation*}
where
\begin{equation*}
  P_{\bar{\rho},\bar{\kappa}}(\mathrm{acc}\mid s_t,\tilde{s},\phi)
  =
  \sigma\!\left(
    \bar{\kappa}
    \left[
      V_\phi(\tilde{s}) - V_\phi(s_t)
      - \bar{\rho}\, d^2(s_t,\tilde{s})
    \right]
  \right).
\end{equation*}
The population-myopic baseline is an evaluability-aware one-step planner with fixed population-level response parameters $(\bar{\rho},\bar{\kappa})$.
It tests whether a generic evaluability model is sufficient, without personalising the user's distance sensitivity or response sharpness from interaction.
Unlike Value-greedy and Threshold, it uses proposal distance through the predicted acceptance probability, so distant proposals can be down-weighted even when their expected value is high.
Unlike \textsc{ProSE-Plan} and the personalised myopic baseline below, it updates only the belief over $\phi$ from observed responses; $\bar{\rho}$ and $\bar{\kappa}$ remain fixed throughout the episode.
The method is also myopic: it scores only the expected value after the current accept/reject response and does not plan for how that response would change future proposals.
For the main corridor experiments, we set $(\bar{\rho},\bar{\kappa})=(0.18,1.0)$.
These values are fixed a priori, not tuned on validation seeds, because the purpose of this baseline is to represent a generic typical-user model rather than the best validation-optimised fixed-response model.
The value $\bar{\rho}=0.18$ is the central value of the synthetic corridor evaluability range used in the main sweep, which spans low-cost users through $\rho_{\mathrm{true}}=0.36$.
Thus, the baseline is deliberately well matched to moderate-cost users and is expected to be competitive near $\rho_{\mathrm{true}}=0.18$.%, while testing whether such a single typical-user setting remains adequate away from that point.
The value $\bar{\kappa}=1.0$ matches the response sharpness used to generate the main corridor data, giving this baseline the correct response-noise scale and isolating the effect of using a non-personalised evaluability slope.

\paragraph{Personalised-myopic.}
\begin{equation*}
\tilde{s}_*
=
\arg\max_{\tilde{s} \in \mathcal{C}(s_t)}
\sum_{z} b_t(z)
\Big[
  P(\mathrm{acc}\mid s_t,\tilde{s},z)
  V_\phi(\tilde{s})
  +
  \big(1-P(\mathrm{acc}\mid s_t,\tilde{s},z)\big)
  V_\phi(s_t)
\Big],
\end{equation*}
where
\begin{equation*}
P(\mathrm{acc}\mid s_t,\tilde{s},z)
=
\sigma(
  \kappa
  [
    V_\phi(\tilde{s}) - V_\phi(s_t)
    - \rho\, d^2(s_t,\tilde{s})
  ]
).
\end{equation*}
The personalised-myopic baseline is the one-step version of the evaluability-aware planner.
It uses the full posterior belief over $z=(\phi,\rho,\kappa)$ to predict acceptance probabilities and to score the immediate accept/reject outcome of each candidate.
Unlike population-myopic, it personalises the evaluability slope and response sharpness from observed responses by updating the posterior over $\rho$ and $\kappa$ as well as $\phi$.
However, the proposal score is still myopic: it includes only the value of the state reached after the current response, and does not include the downstream value of how that response would change future proposals.
This baseline therefore isolates the value of personalised evaluability inference without response-contingent lookahead.

\paragraph{Belief-frozen depth-2.}
\begin{equation*}
  \tilde{s}_*
  =
  \arg\max_{\tilde{s}\in\mathcal C(s_t)}
  \sum_{y\in\{\mathrm{acc},\mathrm{rej}\}}
  p_{b_t}(y\mid s_t,\tilde{s})
  \max_{\tilde{s}'\in\mathcal C(s_t^y)}
  Q_1^{\mathrm{frozen}}(s_t^y,b_t,\tilde{s}'),
  \end{equation*}
  where
  \begin{equation*}
  Q_1^{\mathrm{frozen}}(s,b,\tilde{s}')
  =
  \sum_{y'\in\{\mathrm{acc},\mathrm{rej}\}}
  p_b(y'\mid s,\tilde{s}')
  H(s^{y'},b).
  \end{equation*}
This ablation performs depth-2 lookahead with the same horizon as \textsc{ProSE-Plan}, but it holds the belief fixed at $b_t$ when evaluating the second-step proposal in each hypothetical first-response branch.
Equivalently, it asks: if the first response changed the task state but did not teach the assistant anything about the user, which of the first proposals would look best?
The method still updates the belief after the actually observed response during execution; the ``frozen'' assumption applies only inside the planning tree used to score hypothetical continuations.
This ablation isolates the value of using a hypothetical response as a learning signal before scoring the follow-up proposal, beyond the value of simply looking two steps ahead in the task state.

\paragraph{Oracle depth-2.}
\begin{equation*}
  \tilde{s}_*
  =
  \arg\max_{\tilde{s}\in\mathcal C(s_t)}
  \sum_{y\in\{\mathrm{acc},\mathrm{rej}\}}
  P(y\mid s_t,\tilde{s},z_{\mathrm{true}})
  \max_{\tilde{s}'\in\mathcal C(s_t^y)}
  Q_1(s_t^y,\delta_{z_{\mathrm{true}}},\tilde{s}'),
  \end{equation*}
where $\delta_{z_{\mathrm{true}}}$ is a point-mass belief on the true latent parameters $z_{\mathrm{true}}=(\phi_{\mathrm{true}},\rho_{\mathrm{true}},\kappa_{\mathrm{true}})$.
Oracle depth-2 runs the same depth-2 planner as \textsc{ProSE-Plan}, but removes all user-model uncertainty by giving the assistant the true preference, evaluability slope, and response sharpness.
It is therefore a reference ceiling, which measures how well a depth-2 planner could perform if inference over $z$ were solved perfectly.
Because user responses are still sampled from the stochastic response model, even the true best proposal can be rejected, oracle knowledge does not guarantee success in every finite-horizon episode.

\subsection{Branching-Corridor Additional Results}\label{app:corridor_full}
\subsubsection{Full experimental setup}\label{app:corridor_setup}
\begin{table}[t]
  \centering
  \caption{Branching-Corridor graph, value function, and response-model setup.}
  \label{tab:corridor_setup}
  \small
  \begin{tabular}{lll}
  \toprule
  Component & Notation  & Value  \\
  \midrule
  Branches & $K$ & $4$ \\
  Corridor length & $C$ & $2$ \\
  Branch length & $L$ & $4$ \\
  State space & $\{s_0,c_1,c_2\}\cup\{b_{k,j}\}$ & $k\in\{1,\ldots,4\}$, $j\in\{1,\ldots,4\}$ \\
  Start state & $s_0$ & fixed \\
  Hidden preference & $\phi$ & $1,2,3,4$ \\
  Goal state & $g_\phi$ & $b_{\phi,4}$ \\
  Success & $\mathbf{1}\{s_T=g_\phi\}$ & reach preferred leaf \\
  Horizon & $T$ & $5$ proposals \\
  Candidate set & $\mathcal{C}(s_t)$ & all states except current state \\
  \midrule
  Corridor value scale & $w_c$ & $3.0$ \\
  Preferred branch reward & $w_b$ & $2.0$ per depth \\
  Wrong branch penalty & $w_p$ & $3.0$ per depth \\
  Corridor weight & $\alpha_{\mathrm{env}}$ & main: $0.25$; alignment sweep: $\{0,0.25,0.5,0.75,1\}$ \\
  Value at start & $V_\phi(s_0)$ & $0$ \\
  Corridor value & $V_\phi(c_j)$ & $\alpha_{\mathrm{env}} w_c j$ \\
  Preferred branch value & $V_\phi(b_{\phi,j})$ & $\alpha_{\mathrm{env}} w_c C + w_b j$ \\
  Wrong branch value & $V_\phi(b_{k,j}), k\neq\phi$ & $\alpha_{\mathrm{env}} w_c C - w_p j$ \\
  \midrule
  Distance transform & $g(d)$ & main: $d^2$; robustness: $\{d,d^{1.5},d^2,d^{2.5}\}$ \\
  True evaluability cost & $\rho_{\mathrm{true}}$ & main sweep: $\{0.04,0.08,0.12,0.18,0.24,0.30,0.36\}$ \\
  True inverse temperature & $\kappa_{\mathrm{true}}$ & $1.0$ \\
  Lapse rate & $\epsilon$ & main: $0$; robustness: $\{0,0.05,0.10\}$ \\
  Default belief grid & $\rho$ & $36$ linearly spaced points in $[0.01,0.36]$ \\
  Default belief grid & $\kappa$ & $8$ geometrically spaced points in $[0.5,4.0]$ \\
  Preference prior & $p(\phi)$ & uniform over branches \\
  \bottomrule
  \end{tabular}
\end{table}

This subsection gives the full instantiation of the Branching-Corridor task used in Section~\ref{sec:exp_corridor}.
The environment is a finite tree with one shared corridor and multiple preference-specific branches.
Each episode samples a hidden preferred branch $\phi$, which determines the user's goal state and state-value function.
At each step, the assistant observes the current realised state and may propose any non-current node.
Acceptance moves the state to the proposed node, while rejection leaves the state unchanged.
For belief-based planners, the assistant maintains an exact discrete posterior over $z=(\phi,\rho,\kappa)$ after observing accept/reject responses.
The proposal distance $d(s,\tilde{s})$ is the shortest-path distance on the tree, and the main experiments use the quadratic evaluability transform $g(d)=d^2$.

Table~\ref{tab:corridor_setup} summarises the graph structure, value function, response-model parameters, and sweep ranges.
The default high-cost condition reported in Table~\ref{tab:final_value} uses $\alpha_{\mathrm{env}}=0.25$, $\rho_{\mathrm{true}}=0.30$, and $\kappa_{\mathrm{true}}=1.0$.
Main sweeps use 200 seeds per condition; robustness and ablation checks use 50 seeds unless otherwise stated.

\subsubsection{Additional Branching-Corridor Ablation}
\label{app:corridor_frozen}

The main Branching-Corridor comparison in Section~\ref{sec:exp_corridor} focuses on end-to-end baselines that remove evaluability awareness, personalisation, or non-myopic proposal planning. 
Here we report an additional belief-frozen depth-2 ablation under the same default high-cost condition. 
This ablation has the same depth-2 horizon as \textsc{ProSE-Plan}, but evaluates second-step continuations under the current belief $b_t$, rather than under the posterior induced by hypothetical first-step user responses. 
During actual execution, however, it still updates its belief after observing real user responses.

Table~\ref{tab:corridor_frozen_default} shows that belief-frozen depth-2 outperforms \textsc{ProSE-Plan} in this environment. 
This does not contradict the main Q1 result: the frozen planner is not a non-adaptive baseline, but an internal mechanism ablation that shares the same response model, personalised belief state, and depth-2 horizon. 
Rather, this result indicates that the branching corridor is not a clean isolation test for response-dependent belief updates during lookahead. 
Branch proposals already generate informative real accept/reject outcomes during execution, so a planner can benefit from the resulting posterior updates even if it did not explicitly value hypothetical belief updates inside the lookahead tree. 
In this environment, anticipating the belief update during planning is therefore partly redundant. 
For this reason, we use the Probe-Commit task in Section~\ref{sec:exp_probe_commit} to isolate whether the advantage of \textsc{ProSE-Plan} comes from depth-2 lookahead alone or from response-dependent information probing.

\begin{table}[t]
  \centering
  \caption{
  Default-condition Branching-Corridor ablation including belief-frozen depth-2.
  The setting is $\alpha_{\mathrm{env}}=0.25$, $\rho_{\mathrm{true}}=0.30$, and $\kappa_{\mathrm{true}}=1.0$.
  Values are reported as mean $\pm$ standard error over 200 seeds.
  }
  \label{tab:corridor_frozen_default}
  \small
  \begin{tabular}{lcc}
  \toprule
  Method & $V_\phi(s_T)$ & Success rate \\
  \midrule
  Population-myopic      & $1.90 \pm 0.27$ & $0.200 \pm 0.028$ \\
  Personalised-myopic    & $2.84 \pm 0.28$ & $0.215 \pm 0.029$ \\
  \midrule
  Frozen depth-2         & $\mathbf{7.60} \pm 0.24$ & $\mathbf{0.765} \pm 0.030$ \\
  \textsc{ProSE-Plan}    & $5.62 \pm 0.29$ & $0.515 \pm 0.035$ \\
  \midrule
  Oracle depth-2         & $9.41 \pm 0.07$ & $0.990 \pm 0.007$ \\
  \bottomrule
  \end{tabular}
\end{table}

\begin{table}[t]
  \centering
  \caption{%
  Aggregate proposal behaviour at the default high-cost Branching-Corridor condition
($\alpha_{\mathrm{env}}=0.25$, $\rho_{\mathrm{true}}=0.30$, $\kappa_{\mathrm{true}}=1.0$).
Mean proposal distance and empirical acceptance rate are computed over episodes active at each proposal step.
Values are mean $\pm$ standard error over 200 seeds.}
  \label{tab:corridor_proposal_behaviour}
  \small
  \begin{tabular}{lccccc}
  \toprule
  \multicolumn{6}{c}{Mean proposal distance $d(s_t,\tilde{s}_t)$} \\
  \midrule
  Method & $t=0$ & $t=1$ & $t=2$ & $t=3$ & $t=4$ \\
  \midrule
  Random & $3.94 \pm 0.10$ & $4.00 \pm 0.10$ & $3.75 \pm 0.10$ & $3.73 \pm 0.11$ & $3.70 \pm 0.12$ \\
  Population-myopic & $6.00 \pm 0.00$ & $6.00 \pm 0.00$ & $6.00 \pm 0.00$ & $6.00 \pm 0.00$ & $6.00 \pm 0.00$ \\
  Personalised-myopic & $6.00 \pm 0.00$ & $6.00 \pm 0.00$ & $6.00 \pm 0.00$ & $6.00 \pm 0.00$ & $4.00 \pm 0.00$ \\
  \textsc{ProSE-Plan} & $4.00 \pm 0.00$ & $3.65 \pm 0.05$ & $3.54 \pm 0.07$ & $2.05 \pm 0.02$ & $3.33 \pm 0.11$ \\
  Oracle depth-2 & $3.00 \pm 0.00$ & $3.00 \pm 0.00$ & $3.00 \pm 0.00$ & $3.00 \pm 0.00$ & $3.00 \pm 0.00$ \\
  \midrule
  \multicolumn{6}{c}{Acceptance rate} \\
  \midrule
  Method & $t=0$ & $t=1$ & $t=2$ & $t=3$ & $t=4$ \\
  \midrule
  Random & $0.190 \pm 0.028$ & $0.181 \pm 0.027$ & $0.179 \pm 0.028$ & $0.121 \pm 0.024$ & $0.247 \pm 0.032$ \\
  Population-myopic & $0.035 \pm 0.013$ & $0.047 \pm 0.015$ & $0.043 \pm 0.015$ & $0.034 \pm 0.014$ & $0.059 \pm 0.018$ \\
  Personalised-myopic & $0.050 \pm 0.015$ & $0.074 \pm 0.019$ & $0.051 \pm 0.017$ & $0.060 \pm 0.018$ & $0.185 \pm 0.031$ \\
  \textsc{ProSE-Plan} & $0.175 \pm 0.027$ & $0.330 \pm 0.033$ & $0.355 \pm 0.037$ & $0.662 \pm 0.040$ & $0.355 \pm 0.046$ \\
  Oracle depth-2 & $0.700 \pm 0.032$ & $0.885 \pm 0.023$ & $0.894 \pm 0.038$ & $0.833 \pm 0.090$ & $0.714 \pm 0.184$ \\
  \bottomrule
  \end{tabular}
  \end{table}

\begin{figure}[t]
  \centering
  \includegraphics[width=0.9\columnwidth]{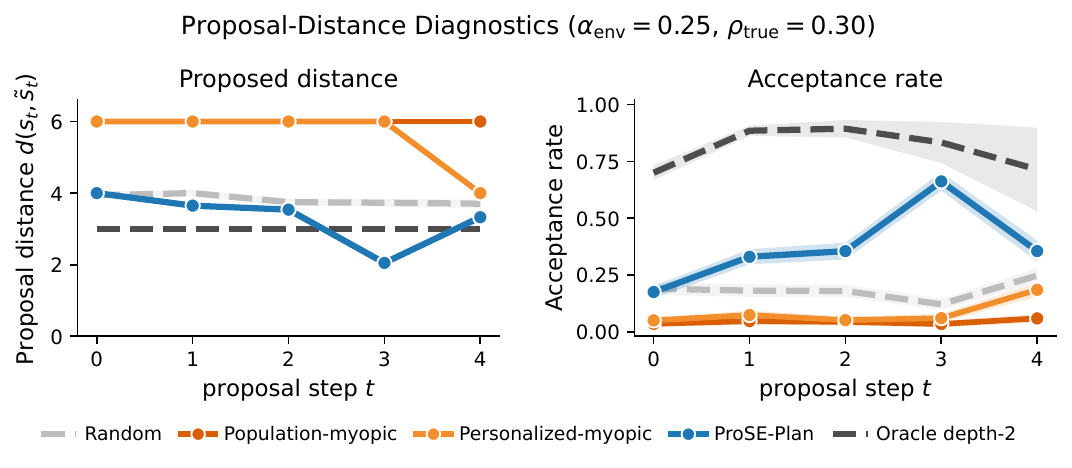}
  \caption{%
  Aggregate proposal behaviour at the default high-cost Branching-Corridor condition.
  Compared with myopic baselines, \textsc{ProSE-Plan} avoids maximal-distance branch jumps and obtains substantially higher acceptance rates, consistent with its use of more evaluable intermediate proposals.}
  \label{fig:corridor_proposal_behaviour}
  \end{figure}
  
\subsubsection{Aggregate Proposal behaviour}
\label{app:corridor_proposal_behaviour}

To understand the behavioural difference behind the Branching-Corridor result, we measure how far each planner proposes from the current state and how often those proposals are accepted. 
For each proposal step $t$, we report the mean proposal distance $d(s_t,\tilde{s}_t)$ and the empirical acceptance rate over episodes that remain active at that step. 

Table~\ref{tab:corridor_proposal_behaviour} and Figure~\ref{fig:corridor_proposal_behaviour} show a clear behavioural separation between \textsc{ProSE-Plan} and the myopic baselines. 
\textsc{Population-myopic} repeatedly proposes maximal-distance branch states with distance $6$, leading to acceptance rates below $0.06$ throughout the episode. 
\textsc{Personalised-myopic} behaves similarly for the first four proposal steps and only shortens its proposal at the final step, after most opportunities for successful progress have already been lost. 
In contrast, \textsc{ProSE-Plan} avoids these maximal-distance jumps: its mean proposal distance decreases from $4.00$ at $t=0$ to $2.05$ at $t=3$, while its acceptance rate increases from $0.175$ to $0.662$. 
At the final step, the mean distance increases again as remaining active episodes require more decisive branch proposals, but the acceptance rate remains substantially above the myopic baselines.

This analysis supports our interpretation of the main result that, \textsc{ProSE-Plan} does not improve merely by proposing lower-value local moves, but by selecting proposals that better balance value gain against evaluability. 

\subsubsection{Representative Proposal Trajectories}
\label{app:corridor_trajectories}
\begin{figure}[t]
  \centering
  \includegraphics[width=0.9\columnwidth]{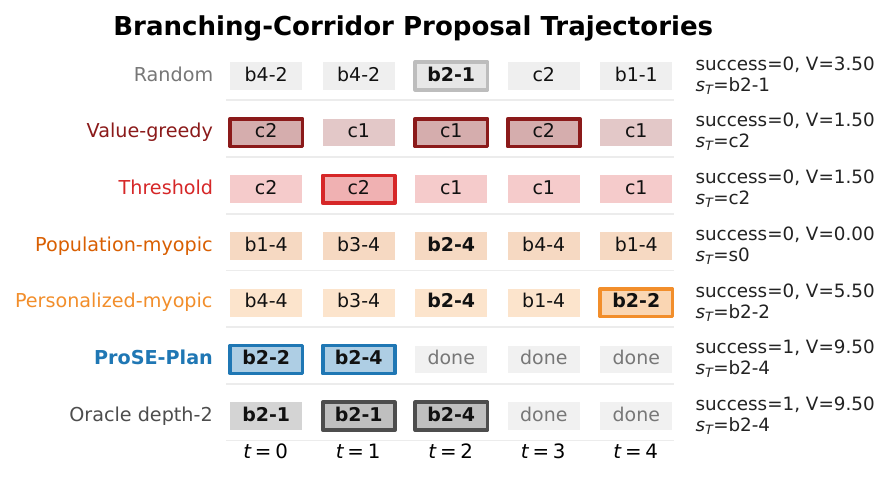}
  \caption{%
  Representative proposal trajectories in the default high-cost Branching-Corridor condition
$(\alpha_{\mathrm{env}}=0.25,\ \rho_{\mathrm{true}}=0.30,\ \kappa_{\mathrm{true}}=1.0)$,
using the same episode seed across all methods (seed $=113$); the true branch is $\mathrm{BR2}$.
Each box shows one proposal.
Dark borders indicate accepted proposals, bold text marks proposals on the true branch, and ``done'' indicates that the episode has already terminated.
The figure qualitatively illustrates three behaviours: conservative corridor proposals (\textsc{Value-greedy}, \textsc{Threshold}), over-ambitious direct branch jumps (\textsc{Population-myopic}, \textsc{Personalised-myopic}), and evaluability-aware stepping-stone proposals (\textsc{ProSE-Plan}).}
  \label{fig:corridor_trajectories}
  \end{figure}

Figure~\ref{fig:corridor_trajectories} shows representative trajectories of all methods in the default high-cost Branching-Corridor condition
$(\alpha_{\mathrm{env}}=0.25,\ \rho_{\mathrm{true}}=0.30,\ \kappa_{\mathrm{true}}=1.0)$, using the same seed for all methods (seed $=113$); the true branch is $\mathrm{BR2}$.
This figure shows that, \textsc{Value-greedy} and \textsc{Threshold} remain in the shared corridor, proposing generic states that are easy to evaluate but insufficient for identifying the preferred branch.
\textsc{Population-myopic} and \textsc{Personalised-myopic} mostly propose distant leaf states directly; under high evaluability cost, these proposals are typically rejected, so progress is limited.
\textsc{ProSE-Plan} instead proposes an intermediate true-branch state $b2\text{-}2$ before committing to $b2\text{-}4$, reaching the preferred goal in two accepted steps.
This example is consistent with our aggregate results, where \textsc{ProSE-Plan} succeeds by using evaluable stepping stones rather than repeated long-distance branch jumps.

\subsection{Probe-Commit Additional Results}
\label{app:isolation}

\subsubsection{Full experimental setup}\label{app:probe_setup}
Table~\ref{tab:probe_commit_setup} summarises the full setup for the Probe-Commit task used in Section~\ref{sec:exp_probe_commit}. 
The task is intentionally minimal: there are two possible user types, two low-value probe states, and two high-value goal states. 
A goal proposal gives high immediate value under both preferences but is weakly diagnostic of the hidden preference $\phi$. 
A probe proposal has lower immediate value, but its accept/reject response is more informative about $\phi$. 
With horizon $T=2$, this structure makes probing useful only if the planner uses the first response to update its belief before choosing the second proposal.
As in the Branching-Corridor task, the assistant may propose any non-current state, proposal distance is the shortest-path distance on the graph, acceptance moves the state to the proposal, and rejection leaves the state unchanged. 
The default condition used in Table~\ref{tab:final_value} sets $w_{p,-}=-3.0$, $\rho_{\mathrm{true}}=0.5$, and $\kappa_{\mathrm{true}}=1.0$. 
Unless otherwise stated, results are averaged over 200 random seeds.

\begin{table}[t]
  \centering
  \caption{Probe-Commit environment setup.}
  \label{tab:probe_commit_setup}
  \small
  \begin{tabular}{lll}
  \toprule
  Component & Notation & Value \\
  \midrule
  Arms & $K$ & $2$ \\
  State space & $\mathcal{S}$ & $\{s_0,p_1,p_2,g_1,g_2\}$ \\
  Start state & $s_0$ & fixed \\
  Hidden preference & $\phi$ & $\{1,2\}$ \\
  Preferred goal & $g_\phi$ & $g_1$ if $\phi=1$, $g_2$ if $\phi=2$ \\
  Success & $\mathbf{1}\{s_T=g_\phi\}$ & reach preferred goal \\
  Horizon & $T$ & $2$  \\
  Candidate set & $\mathcal{C}(s_t)$ & all states except current state \\
  Transition & $s_{t+1}$ & $\tilde{s}_t$ if accepted, $s_t$ if rejected \\
  \midrule
  Value at start & $V_\phi(s_0)$ & $0$ \\
  Matched probe value & $V_\phi(p_\phi)$ & $w_{p,+}=1.0$ \\
  Mismatched probe value & $V_\phi(p_k), k\neq\phi$ & $w_{p,-}=-3.0$ default \\
  Matched goal value & $V_\phi(g_\phi)$ & $w_{g,+}=5.0$ \\
  Mismatched goal value & $V_\phi(g_k), k\neq\phi$ & $w_{g,-}=4.0$ \\
  \midrule
  Distance transform & $g$ & main: $d^2$; linear ablation: $d$ \\
  Default evaluability slope & $\rho_{\mathrm{true}}$ & $0.5$ \\
  Default response sharpness & $\kappa_{\mathrm{true}}$ & $1.0$ \\
  Evaluability range & $\rho_{\mathrm{true}}$ & $\{0.2,0.5,0.8,1.0\}$ \\
  Response-sharpness range & $\kappa_{\mathrm{true}}$ & $\{0.3,0.7,1.0,2.0\}$ \\
  Preference prior & $p(\phi)$ & uniform over two arms \\
  Belief grid & $\rho$ & $21$ linearly spaced points in $[0.01,1.0]$ \\
  Belief grid & $\kappa$ & $8$ geometrically spaced points in $[0.3,4.0]$ \\
  \bottomrule
  \end{tabular}
\end{table}

\begin{figure}[t]
  \centering
  \includegraphics[width=\linewidth]{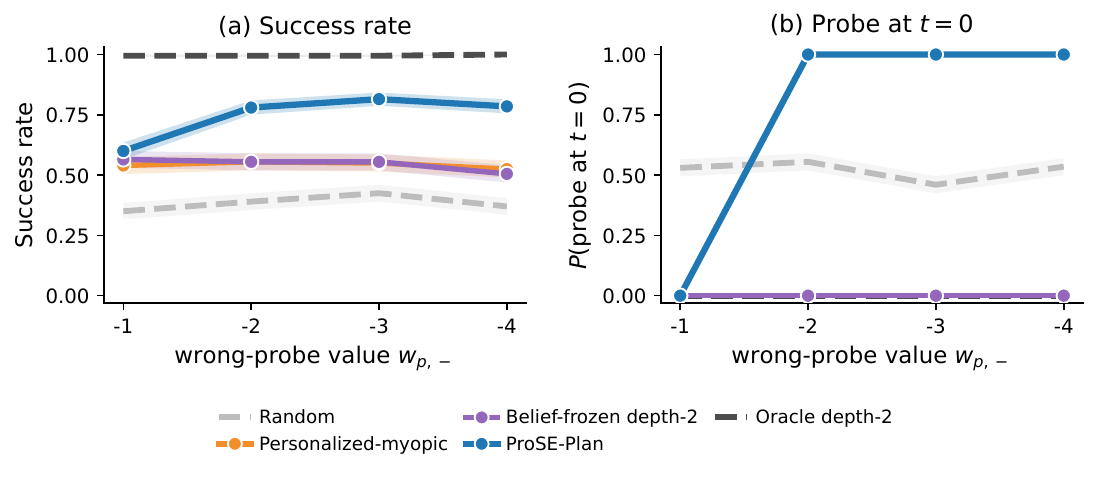}
  \caption{
  Sensitivity to probe diagnosticity in the Probe-Commit task.
  We vary the wrong-probe value $w_{p,-}$, where more negative values make wrong-branch probes easier to reject and therefore more informative about $\phi$.
  \textsc{ProSE-Plan} probes only once probe responses become diagnostic, while myopic and belief-frozen depth-2 planners continue to commit directly.
  }
  \label{fig:probe_diagnosticity}
\end{figure}

\subsubsection{Sensitivity to Probe Diagnosticity}
\label{app:probe_diagnosticity}

We vary the mismatched-probe value $w_{p,-}$, which controls how informative a probe response is about $\phi$.
More negative $w_{p,-}$ makes wrong-branch probes easier to reject, so observing accept/reject at a probe more strongly identifies the preferred branch.

Figure~\ref{fig:probe_diagnosticity} shows that \textsc{ProSE-Plan} probes only in the diagnostic regime.
At $w_{p,-}=-1$, it does not probe and its success is close to the direct-commit baselines.
For $w_{p,-}\le -2$, it probes in every episode and achieves substantially higher success, while \textsc{Personalised-myopic} and \textsc{Belief-frozen depth-2} never probe.
Random probes about half the time but remain much worse, showing that probe selection must be paired with posterior-dependent continuation planning.
We thus conclude that \textsc{ProSE-Plan} does not blindly probe, rather, it probes only when probe responses are sufficiently informative for improving the second proposal.

\begin{table}[t]
\centering
\caption{%
Belief update induced by first-step probes in the Probe-Commit task.
Rows condition on \textsc{ProSE-Plan} proposals at $t=0$; when $w_{p,-}=-1$, \textsc{ProSE-Plan} does not probe, so probe-conditioned quantities are not defined.
Entropy reduction is $H(b_0)-H(b_1)$, and MAP correctness is $\mathbf{1}\{\arg\max_\phi b_1(\phi)=\phi_{\mathrm{true}}\}$.
Values are mean $\pm$ standard error over first-step probe episodes.}
\label{tab:probe_belief_update}
\small
\begin{tabular}{lccccc}
\toprule
$w_{p,-}$ & Probe at $t=0$ & Probe $n$ & $H(b_0)-H(b_1)$ & MAP correct & $b_1(\phi_{\mathrm{true}})$ \\
\midrule
$-1$ & $0.000$ & 0 & --- & --- & --- \\
$-2$ & $1.000$ & 200 & $0.258 \pm 0.011$ & $0.810 \pm 0.028$ & $0.691 \pm 0.017$ \\
$-3$ & $1.000$ & 200 & $0.286 \pm 0.013$ & $0.835 \pm 0.026$ & $0.729 \pm 0.017$ \\
$-4$ & $1.000$ & 200 & $0.310 \pm 0.015$ & $0.810 \pm 0.028$ & $0.729 \pm 0.018$ \\
\bottomrule
\end{tabular}
\end{table}

\subsubsection{Belief Updates Induced by Probes}
\label{app:probe_belief_update}
We further check whether probing actually provides information about the hidden preference. 
For episodes in which \textsc{ProSE-Plan} probes at $t=0$, Table~\ref{tab:probe_belief_update} reports the posterior change after observing the first response. 
The prior over $\phi$ is uniform, so before the first response the assistant assigns probability $0.5$ to the true branch.

When $w_{p,-}=-1$, \textsc{ProSE-Plan} does not probe, so probe-conditioned posterior statistics are undefined. 
For $w_{p,-}\le -2$, the first response to a probe substantially concentrates the posterior,
which confirms that probes selected by \textsc{ProSE-Plan} do provide information that can guide the second-step commit.

\subsubsection{Illustrative Proposal Trajectories}
\label{app:probe_trajectories}
\begin{figure}[t]
  \centering
  \includegraphics[width=0.8\linewidth]{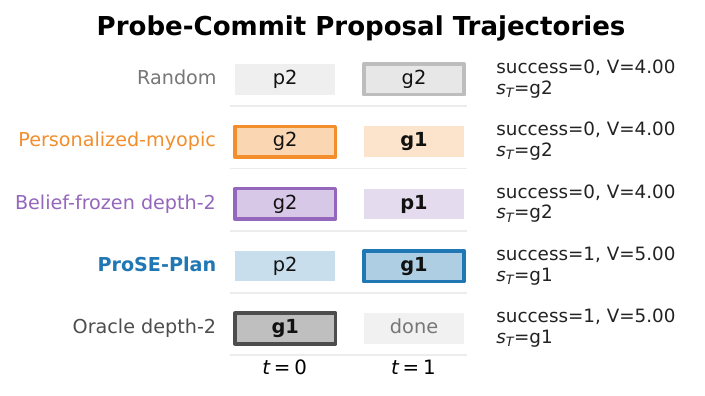}
  \caption{
    Same-seed Probe-Commit proposal trajectories under the default condition
    $(w_{p,-}=-3,\rho_{\mathrm{true}}=0.5,\kappa_{\mathrm{true}}=1.0)$.
    The selected episode uses seed $88$, true branch $\mathrm{BR1}$, and is automatically chosen from seeds matching the main qualitative gap.
    Dark borders indicate accepted proposals; bold labels mark proposals on the preferred branch.
    \textsc{ProSE-Plan} uses a rejected probe to identify the preferred branch before committing, while myopic and belief-frozen planners commit to the wrong goal and fail to recover.
  }
  \label{fig:probe_trajectories}
\end{figure}

Figure~\ref{fig:probe_trajectories} shows selected proposal trajectories under the default Probe-Commit condition
$(w_{p,-}=-3,\rho_{\mathrm{true}}=0.5,\kappa_{\mathrm{true}}=1.0)$ for intuitions on the assistance pattern of each planner.
The trajectories show why direct commitment can fail even when the first proposal is accepted.
Both \textsc{Personalised-myopic} and \textsc{Belief-frozen depth-2} first commit to the wrong goal $g_2$, which can be accepted because the wrong goal still has high value $w_{g,-}=4$.
However, once the realised state moves to $g_2$, correcting to the preferred goal $g_1$ requires a long-distance proposal, and under high evaluation cost, this correction is hard to evaluate and is likely to be rejected, so both planners remain at the wrong goal and fail.

\textsc{ProSE-Plan} behaves differently in that it first proposes the wrong-branch probe $p_2$, which is rejected but informative about $\phi$.
After updating its belief toward $\phi=1$, \textsc{ProSE-Plan} proposes $g_1$, which is now both preference-aligned and reachable from the start state, and the episode succeeds.
The oracle commits directly to $g_1$ because it already knows the preferred branch.
Thus, the trajectory illustrates the mechanism isolated by the Probe-Commit task, which is that the advantage of \textsc{ProSE-Plan} comes from using the first response to change the second proposal, rather than from depth-2 lookahead alone.

\end{document}